\documentclass[journal]{IEEEtran}
\usepackage{graphicx}
\usepackage{cite}
\usepackage{amsmath,amssymb,amsfonts}
\usepackage{algorithm}
\usepackage[noend]{algpseudocode}

\usepackage{textcomp}
\usepackage{url} 
\usepackage{bbm}
\usepackage{calrsfs}
\usepackage{color}
\usepackage{multirow}
\usepackage{tikz}

\usepackage[utf8x]{inputenc}

\usepackage{caption}
\usepackage[framemethod=TikZ]{mdframed}
\usepackage{pgfplots}
\usepackage{colortbl}
\usepackage{xcolor}
\usepackage{comment}
\usepackage[export]{adjustbox}
\definecolor{snm}{rgb}{0,0,0}

\definecolor{dnm}{rgb}{0,0,0}

\usepackage{pgfplots}
\pgfplotsset{compat=1.8}
\usepgfplotslibrary{statistics}

\ifCLASSINFOpdf
\else
\fi
\begin{document}

\title{Interpretability-Driven Sample Selection Using Self Supervised Learning For Disease Classification And Segmentation}

\author{Dwarikanath Mahapatra
\thanks{D. Mahapatra  is with the Inception Institute of Artificial Intelligence, Abu Dhabi, United Arab Emirates. (email: dwarikanath.mahapatra@inceptioniai.org) }% 
}

\maketitle

\begin{abstract}
In supervised learning for medical image analysis, sample selection methodologies are fundamental to attain optimum system performance promptly and with minimal expert interactions (e.g. label querying in an active learning setup). In this paper we propose a novel sample selection methodology based on deep features leveraging information contained in interpretability saliency maps. In the absence of ground truth labels for informative samples, we use a novel self supervised learning based approach for training a classifier that learns to identify the most informative sample in a given batch of images. We demonstrate the benefits of the proposed approach, termed Interpretability-Driven Sample Selection (IDEAL), in an active learning setup aimed at lung disease classification and histopathology image segmentation. 
We analyze three different approaches to determine sample informativeness from interpretability saliency maps: (i) an observational model stemming from findings on previous uncertainty-based sample selection approaches, (ii) a radiomics-based model, and (iii) a novel data-driven self-supervised approach.  We compare IDEAL to other baselines using the publicly available NIH chest X-ray dataset for lung disease classification, and a public histopathology segmentation dataset (GLaS), demonstrating the potential of using interpretability information for sample selection in active learning systems. Results show our proposed self supervised approach outperforms other approaches in selecting informative samples leading to state of the art performance with fewer samples.


\end{abstract}

\section{Introduction}
\label{sec:intro}

Supervised Deep Learning (DL) approaches trained on large datasets have shown state of the art performance \cite{DLRevTMI} on medical image analysis tasks such as classification and segmentation. While DL approaches thrive on large labeled datasets, obtaining them is a challenge due to: 1) limited expert availability; 2) intensive manual effort to curate datasets (i.e. sample labeling process); and 3) paucity of images for specific disease labels leading to class imbalance.

Recent methods have taken different approaches to address the data shortage issue by using data augmentation, semi supervised learning and active learning, to name a few. Although conventional data augmentation methods relying on transformations such as rotations, random cropping, flipping, intensity rescaling, etc., artificially increase dataset size, they do not ensure incorporation of true distribution variability. Improved data augmentation via generative models has been proposed for medical image applications, where realistic synthetic images are used for data augmentation purposes \cite{FuCVPR2018,JinMICCAI2018,Bowles2018,YiMedIA2019generative,Mahapatra_CVPR2020}. While synthetic images leverage a better data augmentation process, these approaches are not designed to streamline the time-consuming data curation process needed to incorporate novel real samples. 
Semi-supervised learning solves the issue of limited expert availability and expert annotation work, as it leverages many unlabeled samples and a few labeled samples to train a classifier \cite{CheplyginaMedIA2019}. However, performance of these methods is known to be dependent upon the quality and informativeness of samples, which is not ensured by semi-supervised learning itself \cite{TajbakhshMedIA2020}. 

Active learning (AL) is an interesting learning paradigm to progressively improve a model's performance. AL systems enable a progressive learning capability, which is ideal in clinical setups where improvements over time, based on user-feedback is desired\footnote{e.g. The FDA organization has even recently mentioned their interest to adapt their regulations to facilitate exploitation of active learning}. However, limited expert availability and required high clinical expertise hampers the annotation of medical images. Hence sample selection methodologies in medical image analysis applications based on supervised learning are fundamental to attain system performance promptly with minimal clinical expert interactions. We refer to this as active sample selection, where most informative unlabeled samples are selected, queried for labels, and subsequently added for further model training \cite{AL1}. Additionally, we remark that active sample selection methodologies leveraging improved learnability of models is particularly important when AL-based technologies are required to swiftly adapt to potential changes of the imaging protocol, vendor type, model, etc. In the next section we review the state of the art in active sample selection.

\section{Prior Work on Active Sample Selection}

Classical approaches for sample selection in AL methods include entropy-based sample selection \cite{EntropyAL,KuanarVC,MahapatraTMI2021,JuJbhi2020,Frontiers2020,Mahapatra_PR2020,ZGe_MTA2019,Behzad_PR2020,Mahapatra_CVIU2019,Mahapatra_CMIG2019,Mahapatra_LME_PR2017}, uncertainty sampling \cite{Lewis94,Zilly_CMIG_2016,Mahapatra_SSLAL_CD_CMPB,Mahapatra_SSLAL_Pro_JMI,Mahapatra_LME_CVIU,LiTMI_2015,MahapatraJDI_Cardiac_FSL,Mahapatra_JSTSP2014,MahapatraTIP_RF2014,MahapatraTBME_Pro2014,MahapatraTMI_CD2013,MahapatraJDICD2013}, query-by-committee (QBC) \cite{Freund97,MahapatraJDIMutCont2013,MahapatraJDIGCSP2013,MahapatraJDIJSGR2013,MahapatraTrack_Book,MahapatraJDISkull2012,MahapatraTIP2012,MahapatraTBME2011,MahapatraEURASIP2010,MahapatraTh2012,MahapatraRegBook} and density weighting \cite{SettlesCraven2008,TongDART20,Mahapatra_MICCAI20,Behzad_MICCAI20,Mahapatra_CVPR2020,Kuanar_ICIP19,Bozorgtabar_ICCV19,Xing_MICCAI19,Mahapatra_ISBI19,MahapatraAL_MICCAI18,Mahapatra_MLMI18,Sedai_OMIA18,Sedai_MLMI18}. These approaches have been used in medical imaging applications such as segmenting anatomical structures \cite{IglesiasActive,MahapatraMICCAI_CD2013,MahapatraGAN_ISBI18,Sedai_MICCAI17,Mahapatra_MICCAI17,Roy_ISBI17,Roy_DICTA16,Tennakoon_OMIA16,Sedai_OMIA16,Mahapatra_OMIA16,Mahapatra_MLMI16,Sedai_EMBC16} and detecting cancerous regions \cite{DoyleBMC11,Mahapatra_EMBC16,Mahapatra_MLMI15_Optic,Mahapatra_MLMI15_Prostate,Mahapatra_OMIA15,MahapatraISBI15_Optic,MahapatraISBI15_JSGR,MahapatraISBI15_CD,KuangAMM14,Mahapatra_ABD2014,Schuffler_ABD2014,Schuffler_ABD2014_2,MahapatraISBI_CD2014}. 

With the advent of deep learning based approaches, active sample selection has been investigated to accelerate learning in deep active learning setups. Different sample selection strategies have been investigated, including sample entropy \cite{AL1,AL_CEAL,Zhu2017generative,MayerASAL,MahapatraMICCAI_CD2013,Schuffler_ABD2013,MahapatraProISBI13,MahapatraRVISBI13,MahapatraWssISBI13,MahapatraCDFssISBI13,MahapatraCDSPIE13,MahapatraABD12,MahapatraMLMI12,MahapatraSTACOM12,VosEMBC,MahapatraGRSPIE12}, model uncertainty \cite{YangAL_MICCAI17,GorrizML4H2017,Gal2017Active,MahapatraMICCAI2018,BozorgtabarCVIU2019,OzdemirDLMIA2018}, Fisher information \cite{SouratiTMI2019}, and clustering \cite{ZhengAAAI2019,MahapatraMiccaiIAHBD11,MahapatraMiccai11,MahapatraMiccai10,MahapatraICIP10,MahapatraICDIP10a,MahapatraICDIP10b,MahapatraMiccai08,MahapatraISBI08,MahapatraICME08,MahapatraICBME08_Retrieve,MahapatraICBME08_Sal,MahapatraSPIE08,MahapatraICIT06}. 
Proposed entropy-based approaches vary in the way the entropy metric is used. In \cite{AL1,Lie_AR,Salad_AR,Stain_AR,DART2020_Ar,CVPR2020_Ar,sZoom_Ar,CVIU_Ar,AMD_OCT,GANReg2_Ar,GANReg1_Ar,PGAN_Ar,Haze_Ar}, given a pool of candidate samples, the approach selects informative samples by a combination of maximal conditional entropy of the label variable given a candidate sample (sample informativeness), and mutual-information-based density estimation of samples (sample representativeness). In \cite{AL_CEAL,Xr_Ar,RegGan_Ar,ISR_Ar,LME_Ar,Misc,Health_p,Pat2,Pat3,Pat4,Pat5,Pat6,Pat7}, sample entropy along with least confidence and margin sampling metrics are proposed in a general framework where least uncertain samples are pseudo-annotated with a trained oracle. In \cite{CAM,Pat8,Pat9,Pat10,Pat11,Pat12,Pat13,Pat14,Pat15,Pat16}, a query synthesis approach was proposed where a Generative Adversarial Network (GAN) synthesizes samples close to the decision boundary, which are then annotated by human experts. Inspired by this work, instead of querying annotations of synthetic samples, in \cite{MayerASAL} a GAN model is used to generate high entropy samples, which are used as proxy to find most similar real samples from a pool of candidates to be annotated by experts.

%uncertainty

Uncertainty-based sample selection approaches are the most popular among different informative sample selection methodologies. In this paradigm, the basic idea is to select samples for which a model is most uncertain, as they contain new information for model training. In \cite{YangAL_MICCAI17}, a two-step sample selection approach was proposed for computer vision applications, where samples are first selected based on an uncertainty estimation derived via model bootstrapping, followed by a second selection based on a maximum set coverage similarity metric to select representative samples. This idea was followed in \cite{OzdemirDLMIA2018}, but instead of using a two-step approach, the authors propose to combine MC dropout uncertainty-based sample selection, and sample representativeness via a borda-count approach. In addition, different from \cite{YangAL_MICCAI17} where a cosine distance was used as maximum set coverage, the authors in \cite{OzdemirDLMIA2018} propose to measure representativeness of samples via an additional loss cost term optimizing maximum entropy of activation layers. From reported results in \cite{OzdemirDLMIA2018}, it is however not clear how feasible it is to find a good balance between the employed cross-entropy loss (loss term for the main task) and the additional representative loss term, since in their experiments the representative loss term is assigned a very small weight.

Originally proposed in \cite{Gal2017Active}, and later adopted by \cite{GorrizML4H2017,MahapatraMICCAI2018,BozorgtabarCVIU2019} test-time Monte-Carlo dropout was used to estimate uncertainty of samples, and select most informative ones for label annotation. The approaches in \cite{MahapatraMICCAI2018} and \cite{BozorgtabarCVIU2019} differ from \cite{Gal2017Active} as they incorporate a conditional GAN based data augmentation to synthesize similar samples to those selected by the uncertainty criteria, in order to further boost the learning rate. Based on Fisher information metric, the authors in \cite{SouratiTMI2019} propose to select samples based on an efficient low-dimensional approximation of the Fisher information metric targeting Convolutional Neural Networks.  The approach however, relies on a pre-selection step based on sample uncertainty estimation, and its performance hence depends on the sensitivity level of such uncertainty-based pre-selection.
In \cite{ZhengAAAI2019} sample selection is based on a representativeness approach where image patches are projected into a latent space (e.g. via a Variational Autoencoder), clustered in the latent space and sorted by their representativeness using a cosine distance maximum set coverage metric, as done in \cite{YangAL_MICCAI17}. 
 
 % 
% Main idea
 In this paper we propose a novel sample selection approach based on information derived from interpretability saliency maps. Development of interpretability methods for deep learning systems emerged from the need to leverage understanding and insights driving a model's predictions \cite{Reyes2020}. For classification tasks, interpretability saliency maps have been proposed to yield levels of pixel attribution for a given queried class label \cite{Simonyan2013,Springenberg2015,CAM,Shrikumar2017}. Interpretability saliency maps have been proposed to enhance interpretability of deep learning models via visualization of image areas driving predictions, and to perform quality assurance \cite{doshivelez2017rigorous}. Different from prior works, in this paper we propose to use interpretability saliency maps in the context of selecting informative samples for active learning. Our proposition is motivated by the observation that in medical images, sample informativeness strongly relates to information about the studied pathology or condition, which in turn is target of saliency maps highlighting image areas driving a model's prediction.

\subsection{Contributions}
In this paper we make the following contributions:
\begin{enumerate}

\item We propose a novel \textbf{I}nterpretability-\textbf{D}riv\textbf{E}n s\textbf{A}mple se\textbf{L}ection (IDEAL) framework. Up to our knowledge this is the first study showing how interpretability saliency maps can be used to leverage active sample selection. 

\item As part of the proposed framework we present results exploring three different popular approaches, featuring different levels of complexity, to design imaging features extracted from interpretability saliency maps, such as hand-crafted features (observational model), radiomics features, and deep features (data-driven model). We motivate their design and analyze their effectiveness in lung disease classification and histopathology image segmentation.

\item We propose a novel end-to-end deep learning approach to extract deep features from saliency maps, which is trained to identify most informative samples in a self-supervised approach. 

\item We demonstrate the added value of the proposed interpretability-driven active sample selection approach by means of comparison to an standard active learning (i.e. no sample selection involved), and a state-of-the-art uncertainty-driven active learning approach, on a public database of lung-disease classification, and a publicly available dataset for histopathology segmentation.
\end{enumerate}

\section{Methods}
\label{sec:met}

\subsection{Intuition behind IDEAL}
\begin{figure*}[t]
\begin{tabular}{c}
\includegraphics[height=6.2cm, width=18cm]{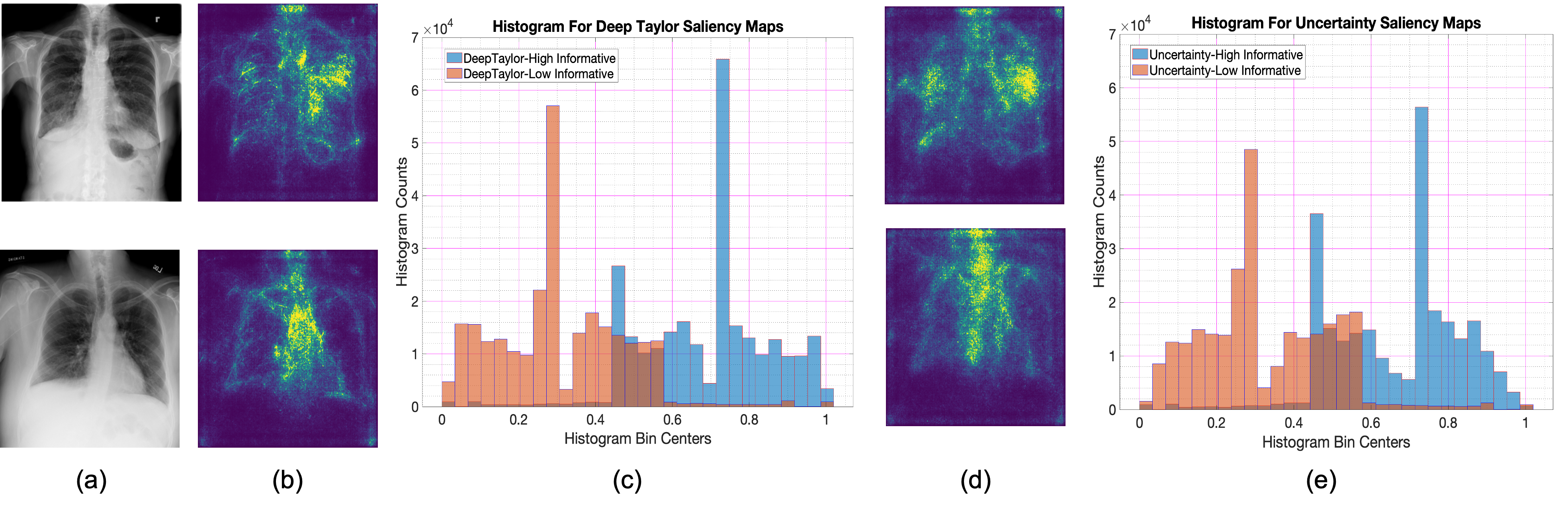} \\
\end{tabular}
\caption{Visualization of saliency maps from different methods. Top row shows a high informative image and bottom row shows a low informative image for pleural effusion condition. (a) original image; (b) saliency map from Deep Taylor method; (c) histograms of saliency  maps for high and low informative images using Deep Taylor; (d) saliency map obtained using Uncertainy; (e) histograms of saliency  maps for high and low informative images using Uncertainty.}
\label{fig:Histograms}
\end{figure*}

In this section we describe the proposed IDEAL approach and its components. We start by presenting the intuition behind using interpretability methods for active sample selection, and the original observation that led us to explore this approach. From our previous studies \cite{MahapatraMICCAI2018,BozorgtabarCVIU2019} using pixel-wise uncertainty maps to perform sample selection we observed a relationship between saliency and uncertainty maps, indicating the possibility of extracting information from these saliency maps to drive sample selection. Figure~\ref{fig:Histograms} shows two example cases of patients with pleural effusion condition, with high and low levels of uncertainty (accumulated pixel-wise uncertainties via MC Dropout). As we compared the corresponding interpretability saliency (derived from Deep Taylor method \cite{alber2019innvestigate}) and uncertainty maps (shown in Fig.~\ref{fig:Histograms}(b) and Fig.~\ref{fig:Histograms}(d), respectively) and their histograms, Fig.~\ref{fig:Histograms}(c) and \ref{fig:Histograms}(e), we observed that the histograms of high and low informative images are quite distinct, thus verifying the fact that high and low informative images have different values for the most salient regions. Comparing between the high informative image histograms of interpretability saliency and uncertainty maps, we observed that the highest peak of the saliency map histogram (i.e. the ``primary'' peak) has a higher count than the corresponding uncertainty map's histogram ``primary peak''. Additionally, the interpretability saliency map histogram ``secondary peaks'' have lower count than those of the uncertainty map histogram. These point to the fact that, compared to the uncertainty method, the interpretability saliency approach identifies salient regions in a more focused manner. This is beneficial when the goal is to identify the most informative saliency maps for improved classification and segmentation.

\begin{figure*}[htbp]
\begin{tabular}{cccccc}
\includegraphics[height=2.2cm, width=2.5cm]{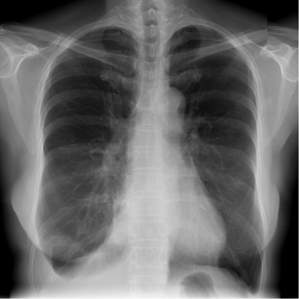} &
\includegraphics[height=2.2cm, width=2.5cm]{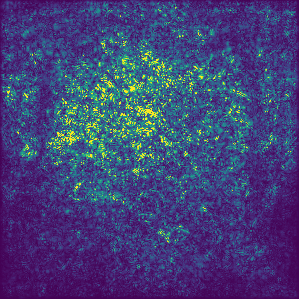} &
\includegraphics[height=2.2cm, width=2.5cm]{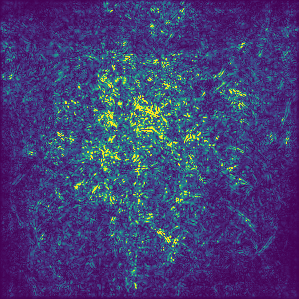} &
\includegraphics[height=2.2cm, width=2.5cm]{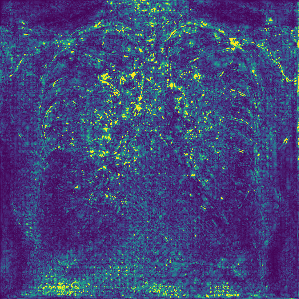} &
\includegraphics[height=2.2cm, width=2.5cm]{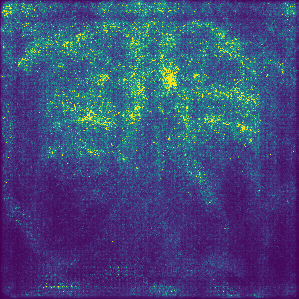} &
\includegraphics[height=2.2cm, width=2.5cm]{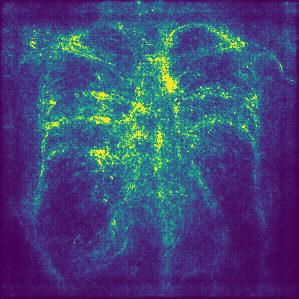} \\
%------------
% &
% ------------
&
\includegraphics[height=2.2cm, width=2.5cm]{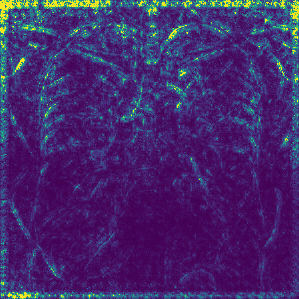} &
\includegraphics[height=2.2cm, width=2.5cm]{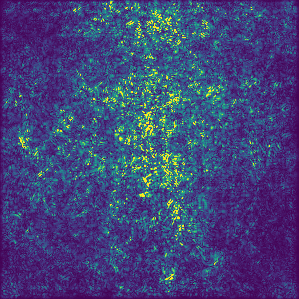} &
\includegraphics[height=2.2cm, width=2.5cm]{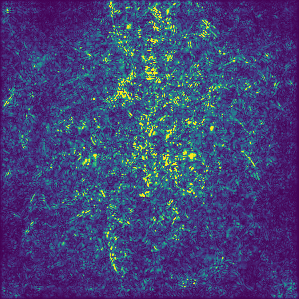} &
\includegraphics[height=2.2cm, width=2.5cm]{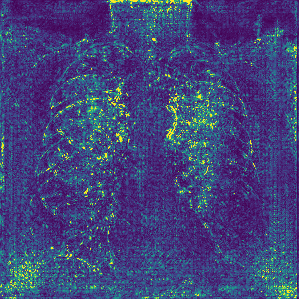} &
\includegraphics[height=2.2cm, width=2.5cm]{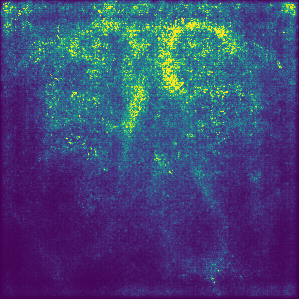} \\
(a) & (b) & (c) & (d) & (e) & (f)\\
\end{tabular}
\caption{Uncertainty maps (top row), and interpretability saliency maps (bottom row) for a given sample input image at different training data levels. (a) Original image (b) $10\%$ of informative samples (c) $20\%$ of informative samples; (d) $40\%$ of informative samples; (e) $60\%$ of informative samples; (f) $90\%$ of informative samples. }
\label{fig:BatchSalMaps}
\end{figure*}

Similarly, we verified how interpretability saliency maps and uncertainty maps vary through the course of training. We selected saliency maps after adding $10\%,20\%,40\%,60\%,90\%$ of training samples. As shown in Figure~\ref{fig:BatchSalMaps}, initially the saliency maps are not well defined since the classifier is not yet trained with sufficient informative samples. However, as more informative samples are being added to the training set, we observe that the saliency maps become more well defined and highlight specific regions of interest. Interestingly, we also observe that saliency maps are sharper and more detailed than uncertainty maps, suggesting an improved description of sample informativeness. The saliency map's informative regions are concentrated and focused on important regions, while the uncertainty maps present dispersed regions. Furthermore at the same percentage of training data, interpretability saliency maps highlight qualitatively better informative regions.

These seminal observations led us to the hypothesis that saliency maps could be used as a proxy to guide an active sample selection. We investigated this hypothesis by investigating how information from saliency maps could be used to guide sample selection. We studied three different approaches to extract information: starting from a simple histogram feature (stemming from our initial observation), followed by a radiomics-based approach, and finally, an end-to-end deep learning based approach cast as a self-supervised learning problem. 
In the following sections we describe the components of the proposed IDEAL approach, including the different investigated information extraction approaches.

\subsection{Main components of IDEAL}

\begin{figure*}[t]
\includegraphics[height=4.2cm,width=18.0cm]{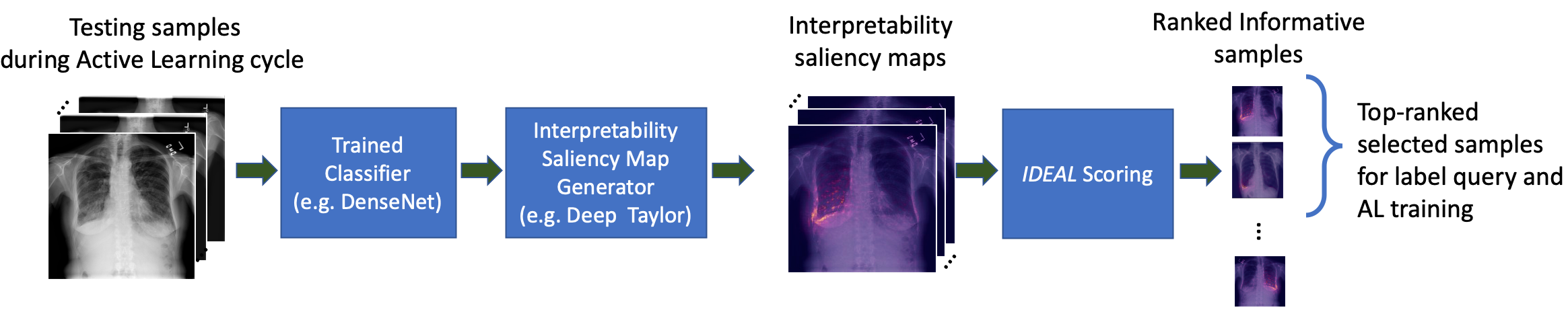} 
\caption{Proposed IDEAL approach. Given testing samples (i.e. pool samples) and a trained classifier, interpretability saliency maps are generated for each testing sample, and an \textit{IDEAL} score is generated from the saliency maps to characterize sample informativeness. Top-ranked samples are then prioritized for label query and for the next active learning training cycle.}
\label{fig:IDEALPipeline}
\end{figure*}

Figure~\ref{fig:IDEALPipeline} depicts a general pipeline of the proposed IDEAL approach. Given unlabeled testing samples (i.e. sample pool), and an associated deep learning classification model (e.g. DenseNet) trained iteratively during active learning, an interpretability saliency map generator is used to produce saliency maps, from which a sample informativeness score (IDEAL Scoring) is calculated to rank pool samples by their informativeness. The IDEAL scoring can be produced in different ways, depending on how the information from the saliency maps is distilled to produce a ranking score for each pool sample. In this study we investigated three different ways of extracting information and scoring samples, which are presented in order of complexity: (i) From our original observation, a single feature extracted from the histogram of saliency maps (Fig.~\ref{fig:Histograms}(c)) is used to derive the IDEAL scoring, (ii) Multivariable radiomics features are extracted and combined into single IDEAL scores, and (iii) Our proposed novel Deep features extracted and used within a self-supervised approach to score informative samples.

In the following we describe each component in detail and in relation to the clinical problem of automating lung disease classification and histopathology segmentation, as well as baseline methods used to benchmark the proposed IDEAL approach.

\subsection{Classification model}

The classification model is not per se a component of the IDEAL approach but rather an input its calculations are based on. We present it here to facilitate the presentations and descriptions of the data workflow, as presented in Fig.~\ref{fig:IDEALPipeline}. Any robust classification model can be used as the approach is not restricted to particular architectures. For lung disease classification from X-ray images, we experimented with $3$ different models namely, DenseNet-121 \cite{CheXNet},ResNet-50 \cite{ResNet} and VGG16 \cite{VGG}, and found the DenseNet-121 architecture to perform the best. We denote as $M$, the DenseNet-121 model, and point the reader to section \ref{sec:impldetails}, for further implementation details of the trained model. For the histopathology image segmentation task we used a DenseNet-121 classifier of the histopathology images (benign vs. malign), as a proxy of informativeness for the main task of image segmentation. This was motivated by multi-task learning where tasks are typically intertwined, and the availability of interpretability approaches for classification tasks. As shown in the results section, we show that this approach is effective in the segmentation task as well.

\subsection{Interpretability Saliency Map Generator}

Image-specific saliency maps operate under the the basic principle of highlighting areas of an image that drive the prediction of a model. The importance of these areas can be obtained by investigating the flow of the gradients of a DL model calculated from the model’s output to the input image, or by analyzing the effect of a pixel (or region) to the output when that pixel (or region) is perturbed. This type of visualization facilitates interpretability of a model but also serves as a confirmatory tool to check that machine-based decisions align with common domain knowledge \cite{Reyes2020}. As mentioned, differently from previous works in interpretability, we aim here to employ saliency maps to perform active sample selection. 
To generate interpretability saliency maps we use the iNNvestigate library \cite{alber2019innvestigate} \footnote{https://github.com/albermax/innvestigate}, which implements several known interpretability approaches. We employ Deep Taylor, a known interpretability approach to generate saliency maps, due to its ability to highlight informative regions while yielding minimal importance to other regions. Deep Taylor operates similarly as other interpretability approaches by decomposing back-propagation gradients, of the studied model, into layer-wise relevance maps of individual cell activations, as a function of a queried input sample and class label (e.g. disease class)\cite{montavon2017explaining}.

\subsection{IDEAL Sample Informativeness Score}
In this section we formalize the definition of the IDEAL sample scoring. Given a test image $I \in R^{m \times n}$, a prediction model $M$ being updated via active learning, and the corresponding saliency map $S(I,M) \in R^{m \times n}$, we map the saliency map $S(I,M)$ into a sample informativeness score, termed IDEAL score as:

\begin{equation}
IDEAL_{score}:f(S(I,M))\in R^{m \times n} \rightarrow R.
\label{eq:IDEALScore}
\end{equation}

The function $f$ can have different forms, depending on the way the information is extracted from the saliency map and converted into an informative sample score. We present results investigating three different approaches, described in further detail below.

The IDEAL scores obtained for the set of testing samples are sorted in decreasing order and the top-n ranked samples are chosen for expert label querying and added to the next active learning cycle. The complete IDEAL process is summarized in Algorithm~\ref{alg1}. In Algorithm~\ref{alg1} the model $M_0$ can be  a pretrained network or, as in our experiments, trained with a small part of the training dataset (e.g. 10\% of training).

\begin{algorithm}
\caption{Interpretability-Driven Sample Selection - IDEAL}\label{alg1}
\begin{algorithmic}[1]
\Require Pretrained model $M_0$, \textit{Saliency map operator} $S(\dot)$, $\mathbb{I}_{validation}$, $AUC_{target}$ 
\State $M \leftarrow M_0$ 
\Repeat
\State $\mathbb{I}_{in} \leftarrow \{I_{in}\}$ \Comment{define set of input testing images}
\State $\mathbb{S}_{in} \leftarrow \{S(\mathbb{I}_{in},M)\}$ \Comment{saliency maps given input set and current model}
\State $\{ scores \}_{in} \leftarrow IDEAL_{score}(\mathbb{S}_{in}) $ \Comment{calculate informativeness scores}
\State $ \mathbb{I}_{sort} \leftarrow sort(\mathbb{I}_{in},\{scores\}_{in})$ \Comment{sort $\mathbb{I}_{in}$ in decreasing order by scores}
\State $\mathbb{I}_{train} \leftarrow \mathbb{I}_{sort}\{i=1,...,top\_n\}$ \label{alg1:topn} \Comment{select top-n ranked samples}
\State $\mathbb{L}_{train} \leftarrow expert\_query(\mathbb{I}_{train})$ \Comment{label querying of selected samples}
\State $M_{new} \leftarrow train(M,\mathbb{I}_{train},\mathbb{L}_{train})$ \Comment{train new model}
\Until{$AUC(M_{new},\mathbb{I}_{validation}) \geq AUC_{target}$} \Comment{Repeat until target AUC is attained}\\
\Return $M_{new}$
\end{algorithmic}
\end{algorithm}

We now describe in detail each of the three studied feature extractor approaches:

%%%%% Single Hand-crafted Feature - Kurtosis %%%%
%%%%%%%%%%%%%%%%%%%%%%%%%%%%%%%%%
\vspace{5pt}
\subsubsection{\textbf{Single Hand-crafted Feature - Kurtosis}}
This first approach is motivated by the observation made from the histograms between high and low uncertainty samples, Figure~\ref{fig:Histograms}. As operator $f$ in Equation~\ref{eq:IDEALScore}, we defined $f=k(H(S(I)))$, with  $H$ and $k$ corresponding to the histogram and kurtosis operators, respectively. Consequently, and based on our observations, informative samples are associated to larger kurtosis values and sorted accordingly to select informative ones. In the results section, this approach is referred to as \textit{Kurtosis}.

%%%%% Multivariate Radiomics Features %%%%
%%%%%%%%%%%%%%%%%%%%%%%%%%%%%%%%%%%%%%%%%%
\vspace{5pt}
\subsubsection{\textbf{Multivariate Radiomics Features}}
As second approach, we used the PyRadiomics package \cite{Pyrad} to extract different radiomics features from the saliency maps. Owing to the large number of potential features we employ a feature selection strategy.  PyRadiomics has 8 feature categories. We trained random forest (RF) classifiers to predict the image's disease label using each feature category. Based on the in-built information gain of the RF model, we identified the $3$ best performing categories as: ``First Order Statistics'' ($19$ features), ``Gray Level Co-occurrence Matrix (GLCM)'' (24 features), and ``Shape Based (2D)'' ($10$ features). The final features for each category is identified by performing an exhaustive search over all possible feature combinations and using it to predict disease labels with a RF classifier. The final features are summarized in Table~\ref{tab:features}.   

% consisting of the following steps:

In order to combine extracted selected radiomics features for ranking, we rank different metrics based on the Borda count, which has been used before for ranking informative samples \cite{OzdemirDLMIA2018}. With Borda count samples are ranked for
each metric, and samples are selected based on the best combined rank as:
\begin{equation}
    i^{*}=\arg \min_i \left(\sum_{m_k} f_{rank} m_k(I_i) \right)
\end{equation}
where $m_k$ denotes the $k^{th}$ pyradiomic feature calculated on image $I_i$.

In the results section, this approach is referred to as \textit{PyRad\_category}, with \textit{category} being one of the following \textit{\{1st-order, GLCM, 2DShape\}}.

\begin{table}[!htbp]
 \begin{center}
 \caption{Description of selected Pyradiomics and Deep Saliency Features.}
\begin{tabular}{|l|l|}
\hline 
{\textbf{Feature Type}} & {\textbf{Comments}} \\ \hline
{Kurtosis} & {Obtained from histogram of the intensity distribution} \\ \hline
{Radiomics-} & {Initial $19$ features. Exhaustive search on $2^{19}-1$} \\ 
{First Order} & {combinations. $4$ best features - `Kurtosis',  `Skewness',} \\
 {} & {`Entropy'  and  `Total  Energy'.} \\ \hline
 {Radiomics-} & {Initial $24$ features. Exhaustive search on $2^{24}-1$ } \\ 
{GLCM} & {combinations. $4$ best features -`Sum Entropy',`Inverse  } \\
 {} & {Difference  Normalized',`Difference  Entropy' and } \\ 
 {} & { ‘Maximal  Correlation  Coefficient’.} \\ \hline
  {Radiomics-} & {Initial $10$ features. Exhaustive search on $2^{10}-1$ } \\ 
{Shape} & {combinations. $3$ best features -`Sphericity’, ‘Spherical  } \\ 
{} & {Disproportion’,  and‘Elongation’} \\ \hline
  {Deep Saliency} & { Ordinal Clustering of latent feature vector} \\ 
{Features} & { followed by Self-Supervised step. Details in text. } \\ \hline
\end{tabular}
\label{tab:features}
\end{center}
\end{table}

%%%%% Deep Saliency Features %%%%
%%%%%%%%%%%%%%%%%%%%%%%%%%%%%%%%%
\vspace{5pt}
\subsubsection{\textbf{Deep Saliency Features: Ordinal Clustering And Self Supervised Learning For Informative Sample Selection}}
\label{met:ordinal}

In this section we present the third and more advanced approach. We propose a novel approach that uses deep features extracted from an autoencoder and self supervised learning based ordinal clustering of informative samples.

The goal in self-supervised learning is to identify a suitable self supervision task (or pretext task) that provides additional knowledge (in the form of network weights) to successfully train a model to solve the main task. Some common pretext tasks include for example, estimating relative position of patches \cite{Doersch}, simulated deformation \cite{TongDART20}, segmentation \cite{Mahapatra_MICCAI20}, aggregation learning \cite{Behzad_MICCAI20}, local context \cite{Pathak}, and colour \cite{ZhangECCV2016}. Additionally, exemplar learning has been proposed as a self-supervised learning strategy \cite{Dosovitskiy} where the task is to classify each data instance into a unique class. 

\begin{figure*}[t]
\begin{tabular}{c}
\includegraphics[height=7.4cm,width=16.0cm]{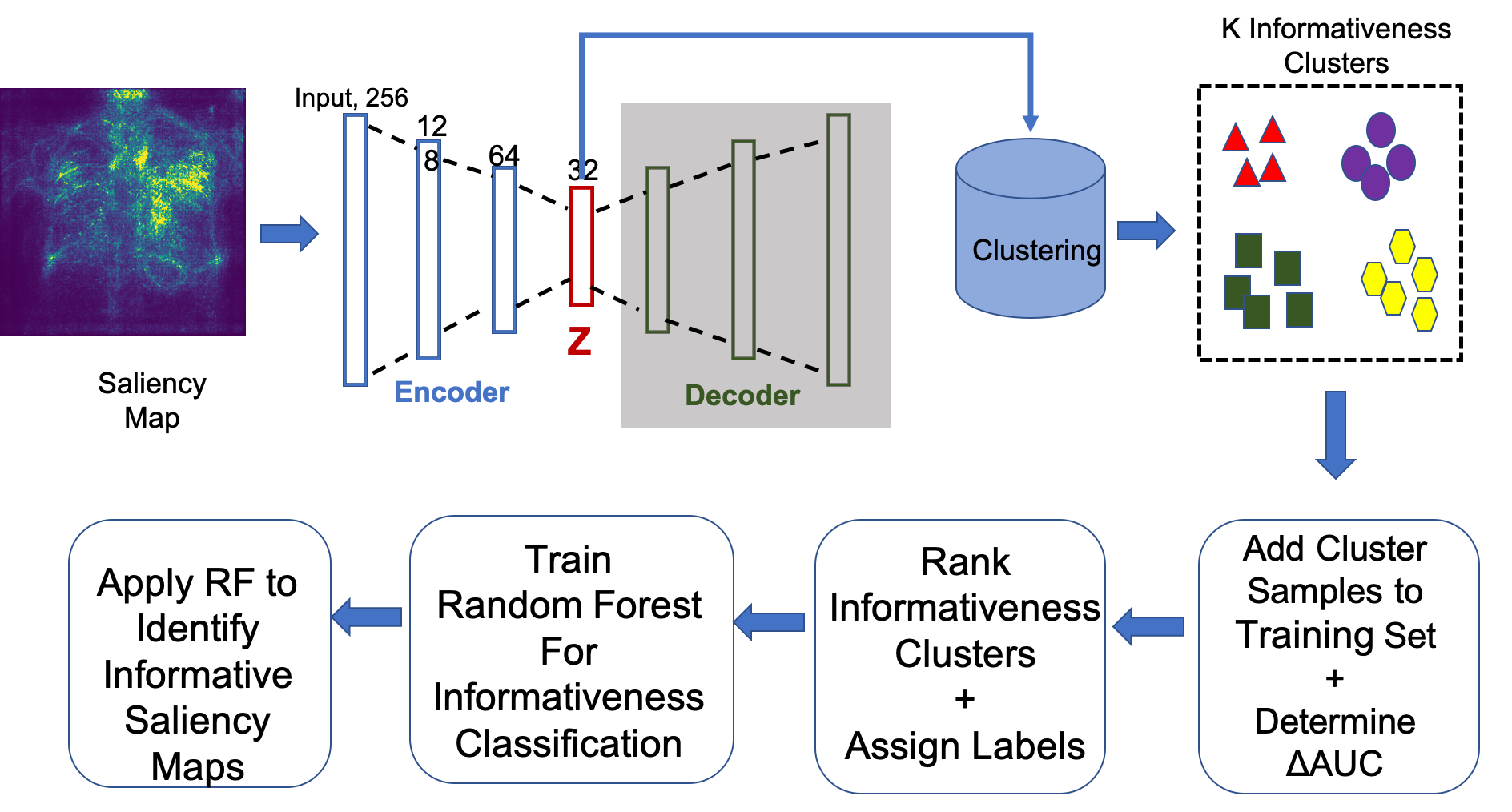} \\
\end{tabular}
\caption{Workflow for Ordinal Clustering and Self Supervised learning for informative sample selection. Deep features extracted from an autoencoder are used for ordinal clustering samples into to $K (e.g. K=10)$ clusters. Representative images from each cluster (i.e. samples closest to each cluster's centroid) are queried for labels and added to the training set. Changes in AUC values on a validation set are determined after updating the classification model. Based on $\Delta$AUC of each cluster's representative image, the $K$ clusters are assigned different levels of informativeness in a self supervised manner. Labels and ranking of representative samples are propagated to samples within each cluster, and random forest is trained on to learn to classify each saliency map into each of $K$-level informative levels. }
\label{fig:SSOrdinal}
\end{figure*}

Given a set of candidate pool samples, saliency maps are generated and, as first step, an auto encoder is trained to reconstruct them. %\
The output of the encoding stage is a $l-$dimensional latent feature vector representation (referred to as \textit{Deep Saliency Features}), which is used as input for the next stage of ranking them. In order to discriminate informative samples from Deep Saliency Features, and in the absence of information to distill which Deep Saliency Features are associated to sample informativeness, we cast the problem as a self supervised learning approach to assign informativeness labels to saliency maps. Figure~\ref{fig:SSOrdinal} depicts the proposed self supervised learning approach, which is also explained in Algorithm~\ref{alg2}. It consists of the following steps:

\begin{enumerate}
    \item Extracted latent feature vector representations are clustered using an ordinal cluster approach into $K(=10)$ clusters. 
    
    \item Identify most representative sample of each cluster via measuring the closest  sample to its centroid using L2 distance between extracted latent feature vectors.
    
    \item Query labels of the most representative sample per cluster (e.g. $K=10$ queries)
    
    \item Add the corresponding original image to the training set, and determine the change in AUC values ($\Delta AUC$) for a fixed validation set (independent of the training set). Rank samples according to decreasing $\Delta AUC$.
    
    \item Identify cluster whose representative image yields maximum positive $\Delta$AUC on validation set.
    \begin{enumerate}
        \item Select this cluster as most informative
        \item Label each cluster as $[1,\cdots,K]$ where $1$ is most informative cluster and $K$ denotes least informative cluster. 
        \item Ranking and queried labels for each representative sample are propagated to all samples within each cluster. 
    \end{enumerate}
    
    \item Use labelled samples from the previous step, and their corresponding deep saliency features to train a random forest classifier. In order to efficiently train the random forest classifier as new samples are selected, we use online random forests \cite{onlineRF}, which performs incremental training of the RF using the previously trained RF as a starting point. The saliency map is classified into one out of $K$ possible levels of informativeness. Use $RF_{final}$ (Algorithm~\ref{alg2}) to rank new (test) samples based on informativeness.

\end{enumerate}

\begin{figure}[t]
\begin{tabular}{c}
\includegraphics[height=4.9cm,width=9.0cm]{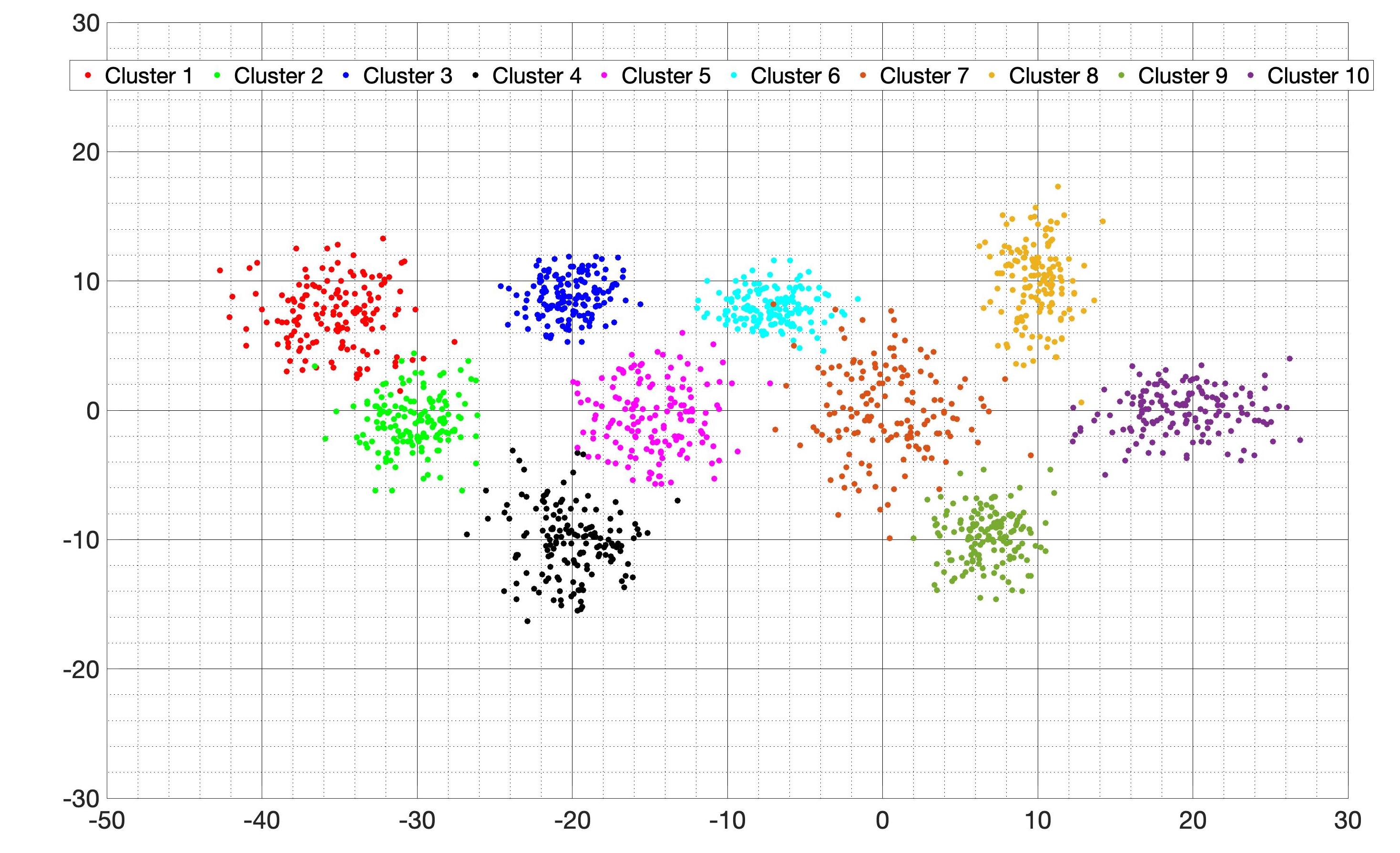} \\
\end{tabular}
\caption{t-SNE plots of different informativeness clusters. `Cluster 1' denotes most informative cluster and `Cluster 10' denotes least informative cluster.} 
\label{fig:ClusterVis}
\end{figure}

Figure~\ref{fig:ClusterVis} shows the t-SNE plots of features from samples belonging to different informativeness clusters, where `Cluster 1' denotes the most informative image cluster while `Cluster 10' denotes the least informative cluster. We see a clear separation between different informativeness clusters. There is some overlap of neighboring clusters, which is due to the similar feature characteristics of similarly informative images/samples.

The choice of $K=10$ was to strike a balance between level of granularity and avoid clusters with too few or no samples. If $K$ is too high then we have to increase the batch size (from 32 to higher) to ensure sufficient samples in each cluster for accurately determining a representative vector. However, increased batch size leads to higher computation cost and poses challenges during training. If $K$ is too low then we lose granularity of informativeness rankings. For example, if $K=5$ then samples with different levels of informativeness will be in one cluster and make it difficult to train a reliable classifier to predict informativeness. Thus $K=10$ gives the best tradeoff between these two considerations.

\begin{algorithm}

\caption{Self Supervised Deep Features - Training Stage  }
\label{alg2}
\begin{algorithmic}[1]

\Require Random forest $RF_0$, Set of Deep Saliency feature vectors $F$, \textit{Ordinal Clustering operator} $C(\dot)$, number of clusters $K$, $\mathbb{I}_{validation}$, $\mathbb{I}_{train}$
\State $RF \leftarrow RF_1 , n=1$ 
\Repeat
\State $\mathbb{F}_{n} \leftarrow \{F_{n}\}$ \Comment{Feature vectors for iteration $n$
\State $\mathbb{C}_{n} \leftarrow \{C(\mathbb{F}_{n})\}$ \Comment{clustering output given input set and clustering operator}}
\State Identify representative samples of each cluster $\mathbb{F}_{rep}$ \Comment{sample closest to each cluster's centroid}
\State Identify corresponding original images $\mathbb{I}_{rep}$
\State Query  label  of  most  representative  sample  per  cluster \Comment{For all K clusters}
\State $\mathbb{L}_{rep} \leftarrow expert\_query(\mathbb{I}_{rep})$ \Comment{label querying of representative samples}
\State $RF_{n} \leftarrow train(RF_{n-1},\mathbb{I}_{rep},\mathbb{L}_{rep})$ \Comment{update pre-trained model using online RF
\State Identify cluster $k$ with highest $+\Delta$AUC
\State Label  each  cluster $[1,\cdots,K]$ \Comment{$1$ is  most informative cluster and $K$ is least informative}}
\State $\mathbb{I}_{train}=\mathbb{I}_{train} \setminus \mathbb{I}_{rep}$  \Comment{Update training set}
\State $n \leftarrow n+1$ \Comment{Go to next iteration}
\Until{$\mathbb{I}_{train}=\emptyset$} \Comment{Repeat until all training samples are used }\\
\Return $RF_{final}$ \Comment{Random forest classifier that ranks samples based on informativeness}
\end{algorithmic}
\end{algorithm}

The proposed ordinal clustering and self supervised learning approach for informative sample selection leverages feature extraction information using modern deep learning technologies. This comes at the cost of a minimal label expert querying of representative samples (i.e. number of cluster $K)$, which based on our experience and the results obtained, yields a good trade-off for clinical utilization. 

In the results section this approach is referred to as \textit{Deep Features}.
In the next section we present results obtained with the proposed and baseline approaches, along with several ablation experiments aiming at leveraging further insights and confirmatory evidence on the benefits of the proposed IDEAL approach.

\section{Baseline Methods For Comparison}
\label{sec:baselines}
In this section we describe the baseline methods used for comparison purposes.

\subsection{Standard Active Learning}
As first baseline we considered a standard active learning framework where no sample selection is considered. In this setup, given a set of testing samples, a subset of samples are randomly chosen for label querying and active learning training. It is worth noting, that in clinical practice the number of samples reflects the amount of user interaction needed to incorporate new samples into the next cycle of active learning, and hence it needs to be kept as low as possible. In the results section we refer to this approach as \textit{Random}. %

\subsection{Uncertainty-driven sample selection}

This corresponds to our second baseline. As proposed in \cite{MahapatraMICCAI2018, BozorgtabarCVIU2019}, uncertainty estimation can be used as a metric of sample informativeness for active learning. %
 Given the deep learning model $M$ used for disease classification, mapping an input image $I$, to a unary output $\widehat{y}\in R$, the predictive uncertainty for pixel $y$ is approximated using:
\begin{equation}
Var(y)\approx \frac{1}{T} \sum_{t=1}^{T} \widehat{y}_t^{2} - \left(\frac{1}{T} \sum_{t=1}^{T} \widehat{y}_t \right)^{2} + \frac{1}{T} \sum_{t=1}^{T} \widehat{\sigma}_t^{2}
\label{eqn:Uncert}
\end{equation}
$\widehat{\sigma}^{2}_t$ is the model's output for the predicted variance for pixel $y_t$, and ${\widehat{y}_t,\widehat{\sigma}^{2}_t}^{T}_{t=1}$ being a set of $T$ sampled outputs.

Similarly as for the other compared approaches, the obtained uncertainty estimates are sorted from high to low uncertainty, and the \textit{top-n} samples are chosen for label querying, and added to the next active learning cycle.  In the results section we refer to this approach as \textit{Uncertainty}.

\subsection{Implementation details}\label{sec:impldetails}

Our method was implemented in TensorFlow. We trained DenseNet-121 \cite{DenseNet} on NIH
ChestXray14 dataset \cite{NIHXray}, and for the histopathology datasets. We used Adam \cite{Adam} with $\beta_1=0.93$, $\beta_2 = 0.999$, batch normalization, binary cross entropy loss, learning rate $1e-4$, $10^{5}$ update iterations and early stopping based on the validation accuracy. % 
The architecture and trained parameters were kept constant across compared approaches. Training and test was performed on a NVIDIA Titan X GPU having $12$ GB RAM.
Images are fed into the network with size $320 \times 320$ pixels. %

We employed 4-fold data augmentation  (i.e. each sample augmented 4 times) using simple random  combinations of rotations ($[-25,25]^{\circ}$), translations ($[-10,10]$ pixels in horizontal and vertical directions), and isotropic scaling ($[0.95-1.05]$ scaling factors).
For generation of interpretability saliency maps, we used default parameters of the iNNvestigate implementation of Deep Taylor \cite{alber2019innvestigate}. 
For uncertainty estimation we used a total of $T=20$ dropout samples with dropout distributed across all layers \cite{BDNN}. 
During active learning the batch size for our experiments was set to $16$. %

As shown in Figure~\ref{fig:SSOrdinal} the encoder stage has 3 layers of $256,128,64$ neurons. The output is a $32$ dimensional latent feature vector which is fed to the decoding stage. The output is supposed to reconstruct the original input using the mean square error loss.

% this section has the experimental results 

\section{Results and Discussion}
\label{sec:results}

\subsection{Dataset Description}\label{sec:data}

As use-case applications we applied and analyzed the proposed IDEAL method and baseline approaches on two tasks: Lung disease classification and histopathology segmentation. \textbf{Classification Dataset:} For lung disease classification we adopted the NIH ChestXray14 dataset \cite{NIHXray} having $112,120$ expert-annotated frontal-view X-rays from $30,805$ unique patients.

\textbf{Segmentation Dataset:} For histopathology segmentation we used the public GLAS digital histopathology image dataset \cite{GlasReview} that has manual segmentation maps of glands in $165$ $H\&E$ stained images derived from $16$ histological sections from different patients with stage $T3$ or $T4$ colorectal adenocarcinoma. The slides were digitized with a Zeiss MIRAX MIDI Slide Scanner having pixel resolution of $0.465\mu$m. The WSIs were rescaled to a pixel resolution of $0.620\mu$m (equivalent to $20\times$ magnification).
  $52$ visual fields from malignant and benign areas from the WSIs were selected to cover a wide variety of tissues. An expert pathologist graded each visual field as either ‘benign’ or ‘malignant’. Further details of the dataset can be found in \cite{GlasReview}.

For the lung classification task, we chose pleural effusion as target condition of a classification model, since it is clinically well defined and among the most important lung disease conditions requiring effective computer assisted diagnosis solutions. 
 We selected $1036$ patient images from the NIH dataset with pleural effusion, which are free of artefacts (e.g. incomplete lung regions) and which can lead to accurate visual assessment by an expert radiologist.
 As sanity check, we additionally asked our expert chest radiologist to inspect and confirm the training and testing cases, corresponding labels, as well as the corresponding interpretability saliency maps. 
 
 For each task, the dataset was split into training ($70\%$), validation ($10\%$) and test ($20\%$), at the patient level such that all images from one patient are in a single fold. 

\subsection{Results for IDEAL and Baseline approaches}

In this section we present the main results obtained by the proposed IDEAL approach, and the baselines described in section \ref{sec:baselines}. As evaluation metrics we adopted the Area Under the Curve (AUC), as typically done for active sample selection studies in classification tasks, and the Dice coefficient for the segmentation task. For both tasks, we assessed the metrics for different methods at every $10\%$ increment of training data. Note that we perform the training from scratch using data augmentation and do not use any pre-trained network weights.

\begin{figure}[t]
\begin{tabular}{c}
\includegraphics[height=6.4cm, width=8.7cm]{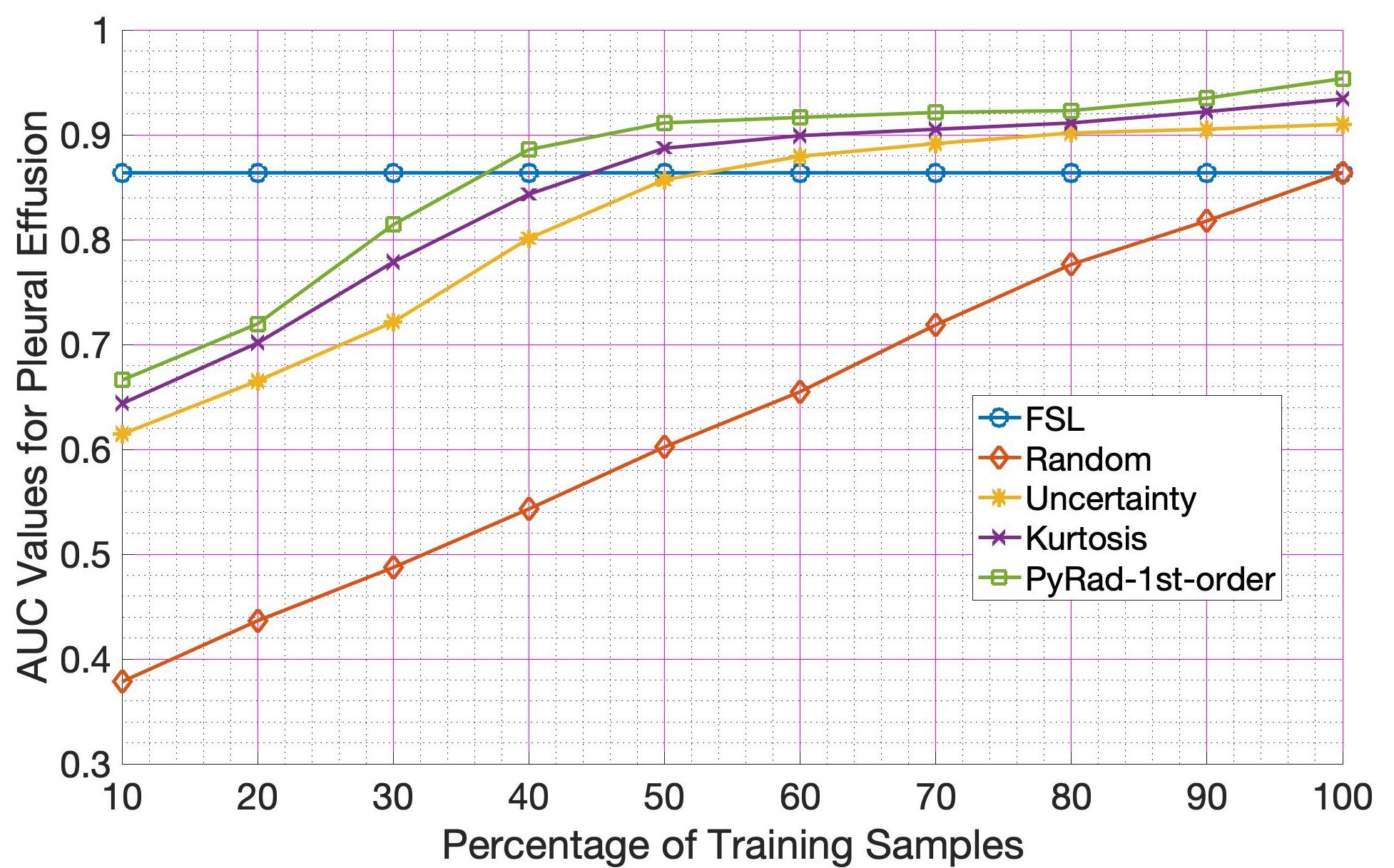} \\
(a) \\
\includegraphics[height=6.4cm, width=8.7cm]{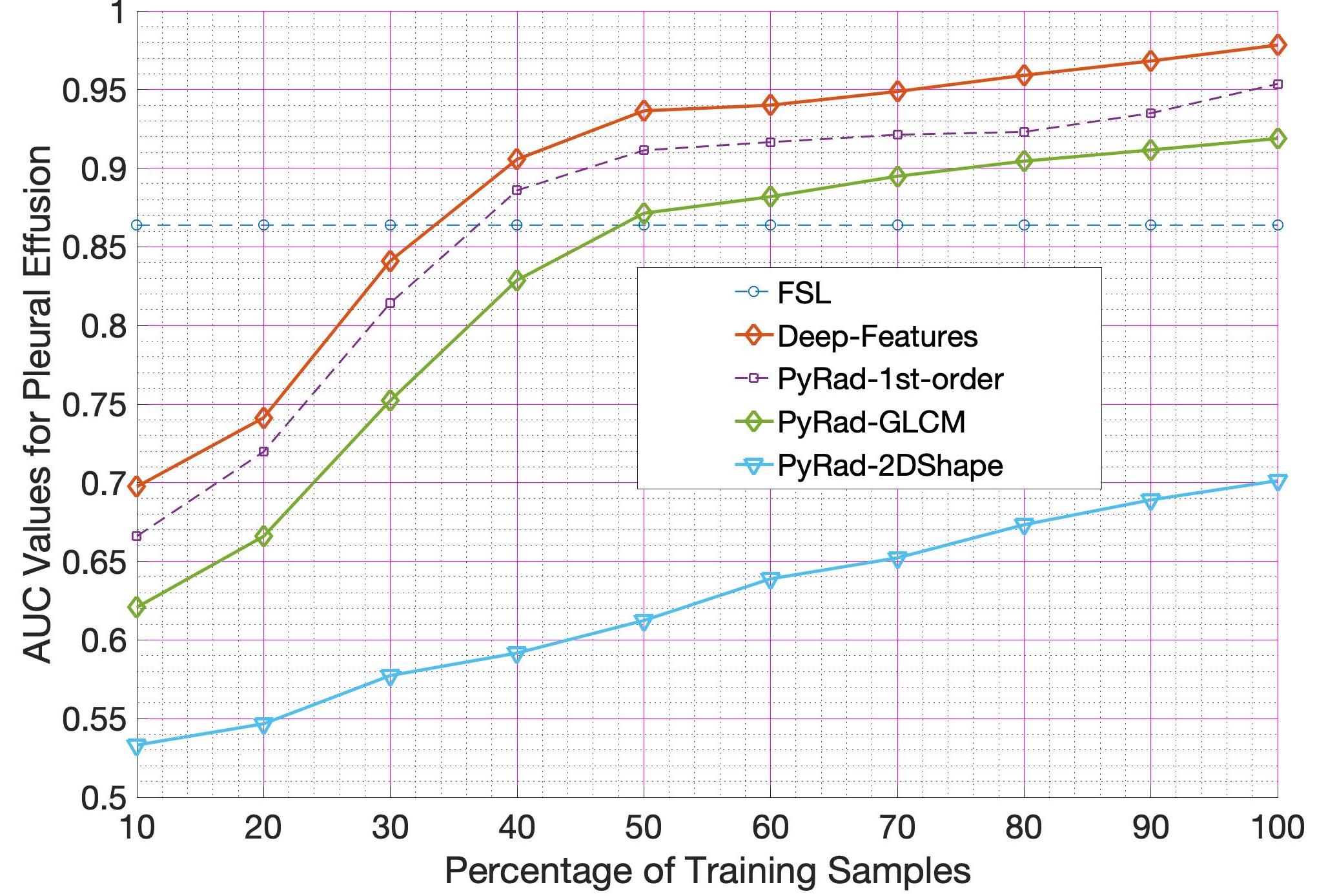} \\
(b) \\
\end{tabular}
\caption{AUC measures at different percentage levels of  training percentage for baselines and proposed IDEAL approach. (a) Baselines (Random and Uncertainty)), IDEAL: Kurtosis and Radiomics based features for active sample selection; (b) IDEAL: Deep Saliency Features with proposed self-supervised ordinal clustering. As reference, in (b) we show best results from (a) using Radiomics 1st-order (PyRad-1st-order). Additionally, other radiomics-based results yielded via GLCM texture features, and 2D-shape based features are presented for completeness. As reference, AUC of a fully-supervised model (FSL) is also included as an horizontal line.}
\label{fig:ALcurves}
\end{figure}

For readability purposes, we split the presentation of our main results in Figure~\ref{fig:ALcurves} in two plots: Figure~\ref{fig:ALcurves}(a) shows results for baselines (Uncertainty and Random), and approaches of Kurtosis and best Radiomics (PyRad-1st-order), and  Figure~\ref{fig:ALcurves}(b), where we present results for Deep Saliency Features, and the best result from Figure~\ref{fig:ALcurves}(a) using 
Radiomics (PyRad-1st-order), which is shown as a dotted line to provide a performance comparison reference with Figure~\ref{fig:ALcurves}(a).

For completeness, we additionally include in Fig.~\ref{fig:ALcurves}(b) other radiomics-based results yielded via GLCM texture features, and 2D-shape based features. 
On both plots in Fig.~\ref{fig:ALcurves}, results with a fully-supervised model (FSL, AUC=$0.8686$)) are also included (horizontal lines in Fig.~\ref{fig:ALcurves}(a) and (b)).  

 From Fig.~\ref{fig:ALcurves}(a) we observe that, except for Random-based sample selection, all approaches outperform the fully-supervised learning model. Moreover, IDEAL approaches based on Kurtosis and Radiomics outperform uncertainty-based sample selection. Additionally, the uncertainty-based approach required $53\%$ of the training data to surpass FSL, which was surpassed at a much lower value for IDEAL approaches: Kurtosis: $44\%$, and PyRad: $37\%$. We remark that this finding aligns with other similar reports, \cite{MayerASAL,YangAL_MICCAI17,SouratiTMI2019}, but its exploration goes beyond the scope of this study. 
 
 Amongst the different Radiomics based features, best AUC was attained with First-Order features, while GLCM performed poorly, followed by 2D-Shape. We attribute this lower performance of GLCM and 2D-shape based features to a combination of an absence of rich texture and shape information of saliency maps, as well as the difficulty to reliably extract shape parameters with pyradiomics (e.g. finding an appropriate parameterization of threshold values to binarize saliency maps before extracting shape information). 
 
As shown in Fig.~\ref{fig:ALcurves}(b), the third approach, based on Deep Saliency Features in combination with the proposed self-supervised ordinal clustering, yielded the best results, outperforming all other approaches in terms of learning rate and final attainable accuracy. Using IDEAL with Deep Saliency Features, enables the approach to attain same performance as the fully-supervised model at only $33\%$ (versus $53\%$ using uncertainty-based sample selection), with best final performance at $95\%$.
We highlight these results in light of the importance of minimizing expert annotations in the clinical routine while targeting high accuracy levels.

The AUC values were derived from an average of $10$ runs and the statistical significance with respect to Deep Saliency Features was calculated using a paired $t-$test. The final AUC values and the corresponding $p-$values for different methods are as follows: 1) FSL-$0.8638,p=0.0008$; 2) `Deep Saliency Features'-$0.9783$; 3) Kurtosis- $0.0.9341,p=0.01$; 4) Uncertainty -$0.9101,p=0.005$; 5) PyRad-1st-Order - $0.9536,p=0.02$; 6) PyRad-GLCM -$0.9189,p=0.009$; 7) PyRad-2DShape- $0.7013,p=0.0001$.

\subsection{Relationship Between Batch Size And Interaction Cost}

\begin{figure}[t]
\begin{tabular}{c}
\includegraphics[height=5.2cm, width=8.49cm]{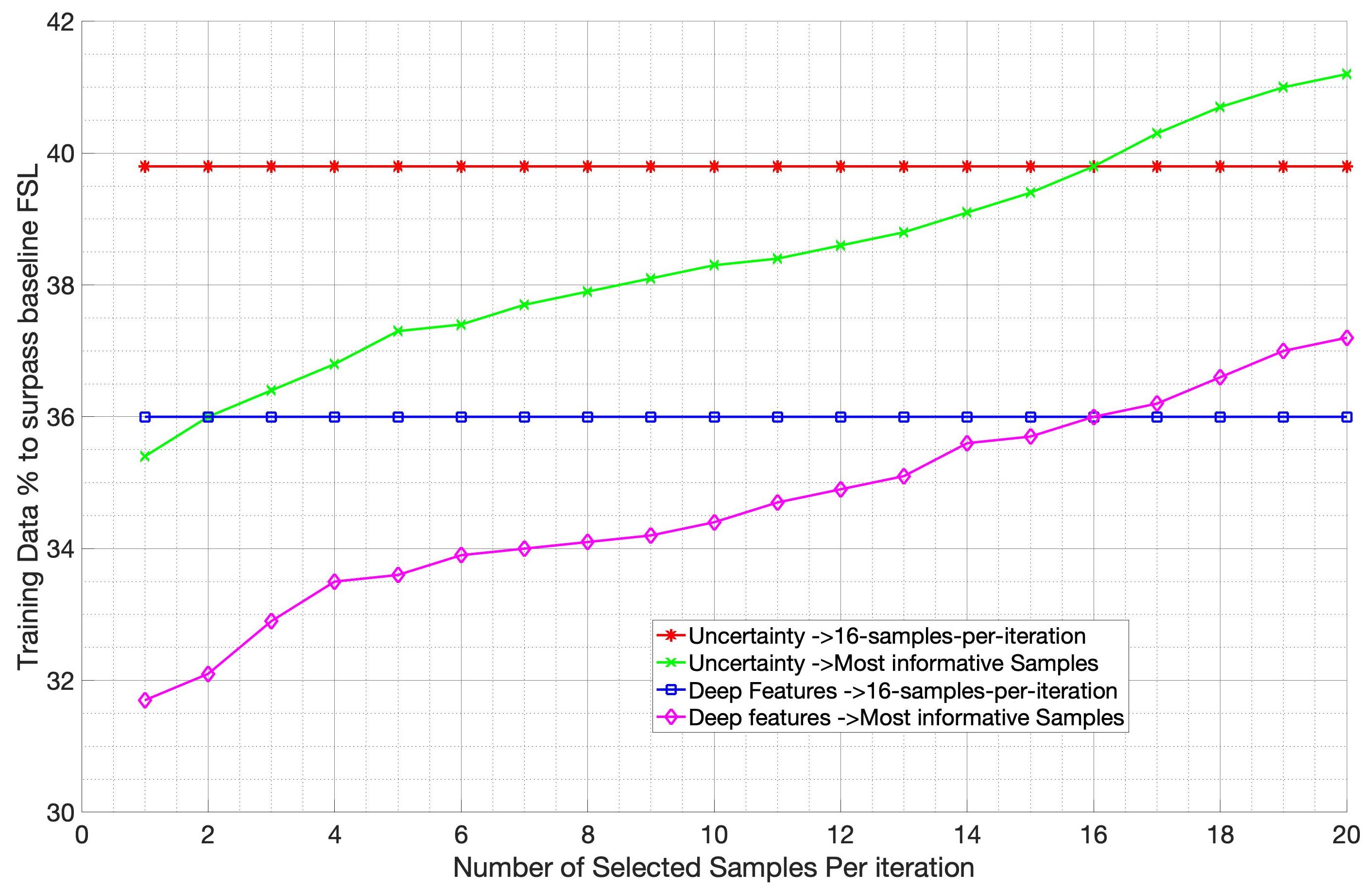} \\
(a) \\
\includegraphics[height=5.2cm, width=8.49cm]{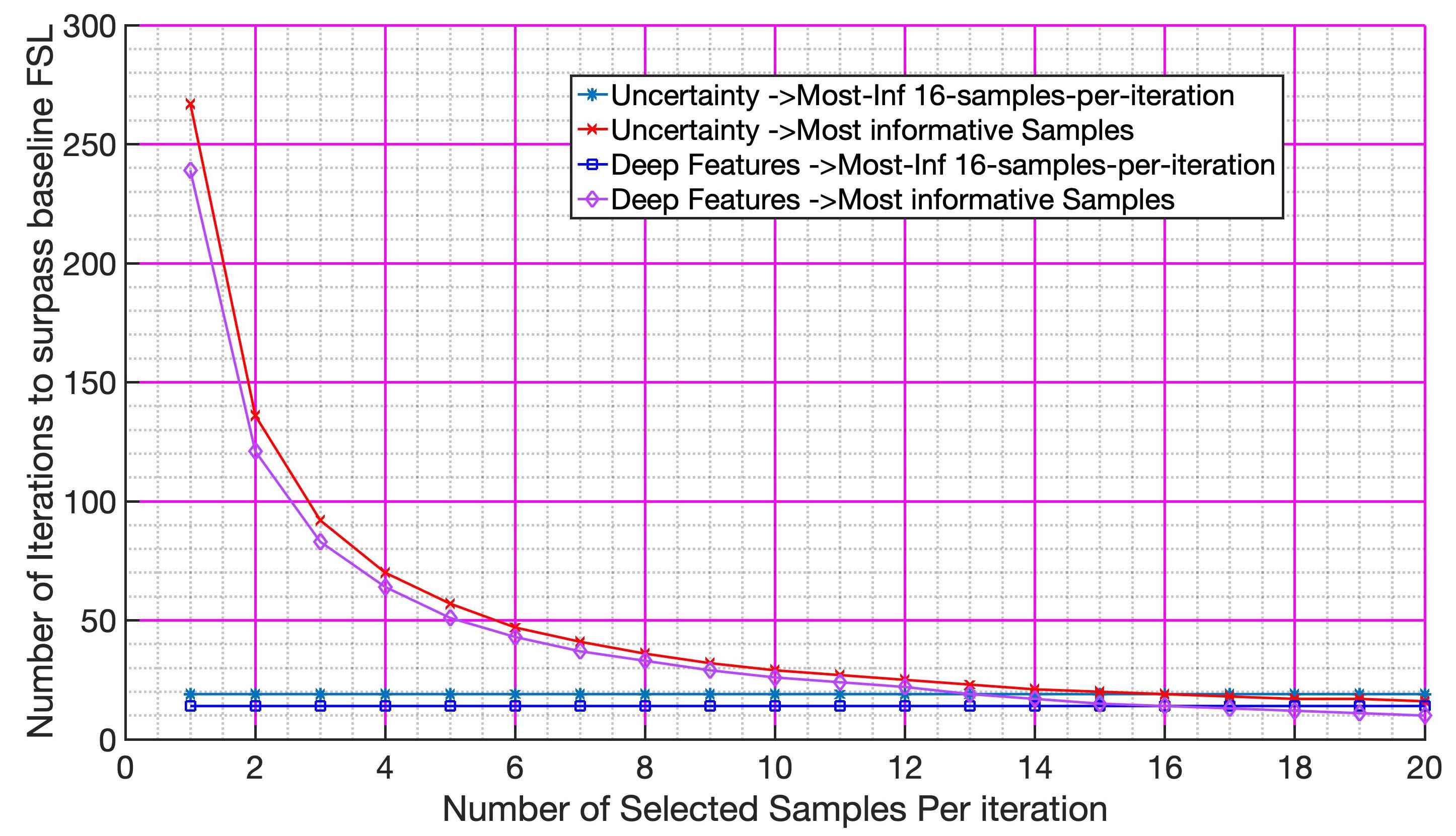}  \\
(b)\\
\end{tabular}
\caption{Interplay between number of selected samples in each iteration and performance: (a) percentage of total training data required to surpass the fully-supervised model ($FSL$); (b) number of iterations required to surpass the fully-supervised model ($FSL$) for varying number of selected samples per iteration.} 
\label{fig:plot2}
\end{figure}

 %-----------------------------------------------
We analyzed the interplay between varying the number of queried samples (line \ref{alg1:topn} in Algorithm \ref{alg1})), and the total number of training iterations required for the active learning system. For this experiment we used the classification task. As reference we used the performance of the fully-supervised model and measured the number of queried samples and training iterations needed by IDEAL to surpass the performance of the fully-supervised model. Figure~\ref{fig:plot2} (a) shows that with reduced number of samples per batch the system can surpass FSL with fewer samples. During the initial phase of fixed-batch size training, queried samples within a single batch might indeed be assessed as being informative, but redundant as they share similar characteristics, leading to the effect that a higher percentage of training dataset is needed to surpass the performance of the fully-supervised model. In the extreme case of selecting one sample per iteration the system will always choose the most informative sample and there will be no redundancy in subsequent sample selections.  

However, as shown in Figure~\ref{fig:plot2} (b), for both IDEAL and uncertainty-based sample selection approaches, a reduction in the number of selected samples per iteration comes at the cost of an exponential increase in the number of training iterations. However, IDEAL requires fewer iterations (on average 10 fewer iterations) than an uncertainty-based sample selection approach.

 This also connects with the phenomenon shown in Figure~\ref{fig:ALcurves} where the baselines based on sample informativeness outperform the fully-supervised model at lower number of training samples. This observation can also be linked to the known phenomenon of influential observations, where samples (observations) have greater influence during the initial training stages than in later ones. We remark that this phenomenon has been reported by others \cite{MayerASAL,SouratiTMI2019,YangAL_MICCAI17}.
 
 Although fewer queried samples per batch can lead to lower percentages of training dataset needed to attain a given performance (compared to using larger batches), there is the higher cost of successive retraining of the model. This can be computationally prohibitive depending on the model size, available resources, etc. For the studied use-case, we found that a batch size of $16$ queried samples per iteration could provide in practice a good trade-off between queried expert labeling and model retraining.
 
 An interesting strategy for sample selection could be that of selecting fewer samples per iteration in the initial learning stages, to then increase the number of selected samples when the classifier has reached a certain performance level. This can potentially ensure that the classifier observes diverse samples in the initial stages while being computationally efficient to reach optimal performance.
 
% 
 %----------------

%----------------------------------

%

%------------------

\subsection{Ablation Studies }
 We performed two ablation experiments to (i) analyze the effect of choosing the least informative samples (instead of the most informative) on the learning curves, and (ii) utilize the input images, instead of the interpretability saliency maps for feature extraction. For these experiments we used the lung disease classification task. 

\begin{figure}[t]
\begin{tabular}{c}
\includegraphics[height=6cm, width=8.7cm]{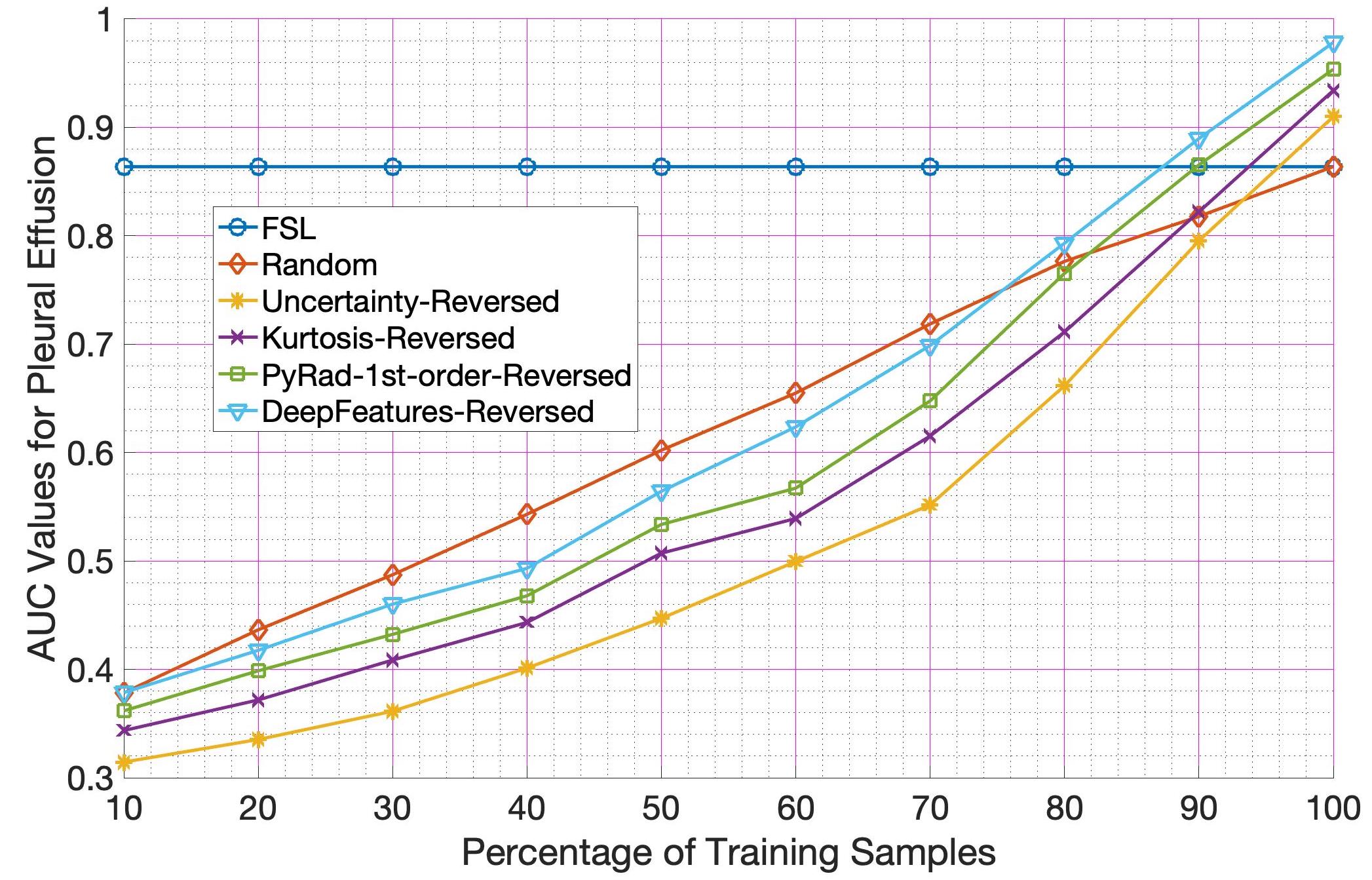} \\
(a) \\
\includegraphics[height=6cm, width=8.7cm]{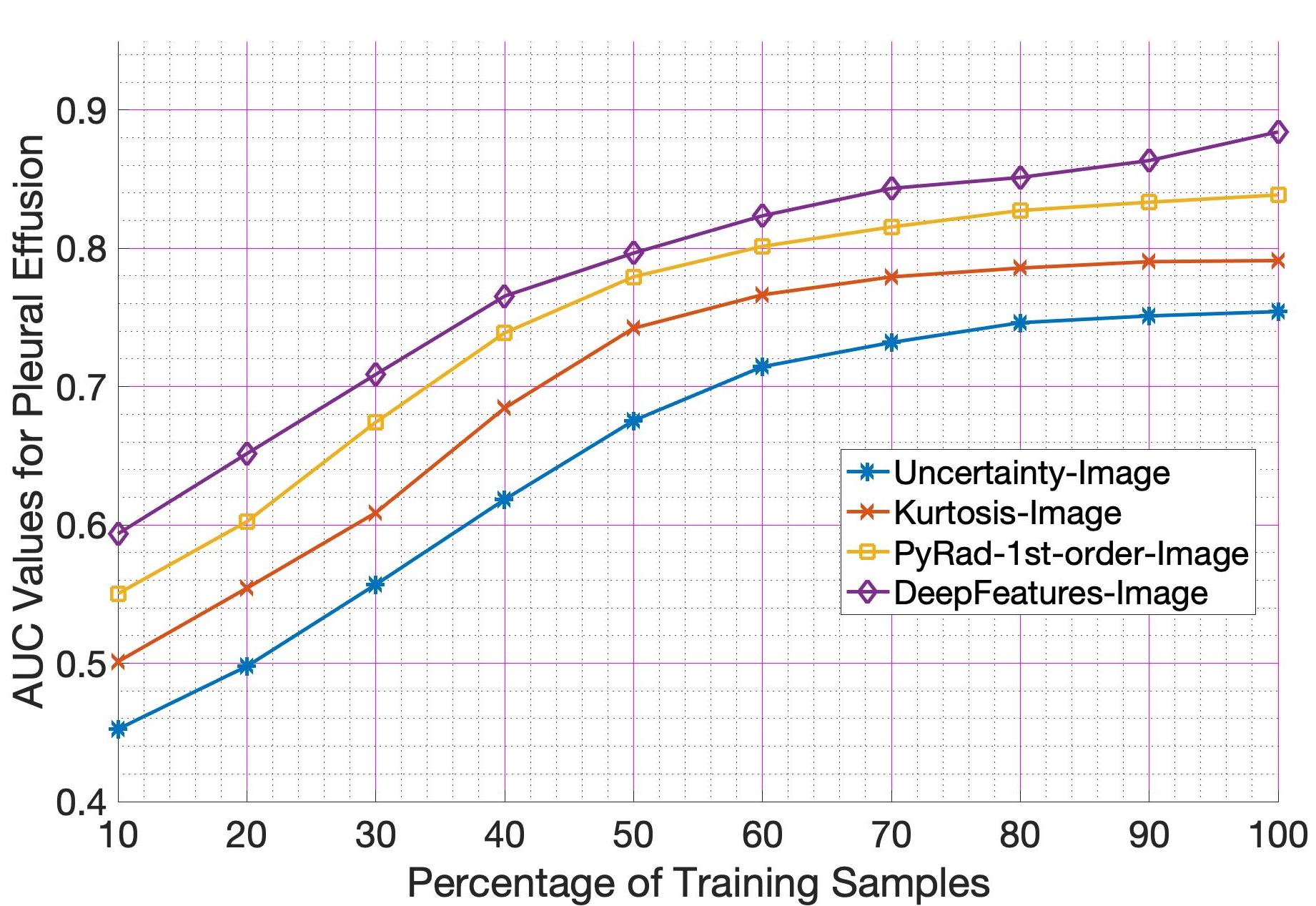}  \\
(b)\\
\end{tabular}
\caption{Ablation studies on the lung classification task to (a) analyze the effect of choosing the least (marked as ``-Reversed)''informative samples (instead of the most informative) on the learning curves, and (b) effect in extracting features from the input images (marked as ``-Image''), instead from the interpretability saliency maps.}
\label{fig:ALRev}
\end{figure}

 For the first ablation experiment, Figure~\ref{fig:ALRev} (a) shows the classification performance when using the least informative samples ( marked as ``-Reversed''). As expected, we observe a very slow increase in the learning rate. Nevertheless, the least informative ``Deep Features'' select better quality features and hence outperform other methods. Similarly, we observed that around $60\%$ of training, the learning rate increases, since the remaining samples are actually the most informative samples in the dataset. This experiment confirms the importance of selecting informative samples.
 
%  For the second ablation experiment,
 Figure~\ref{fig:ALRev} (b) shows the AUC curves when applying the different IDEAL based methods on the original images instead of the saliency maps (marked as ``-Image''). The performance of each approach is lower than the corresponding one when extracting features from the saliency maps. We attribute this to our intuition that saliency maps highlight information regarding the pathology, which is in turn the target of the classification model being explained. In contrast, the X-ray image includes other sources of information, including the overall anatomy, that is of much lower relevance for the trained model.

\subsection{Sample Selection Performance when Switching Datasets}

We assessed the proposed approach on a scenario where, after initial training with one dataset, the sample selection method may be used on a different new dataset in a clinical scenario (e.g. change of imaging vendor). 
To simulate this situation we used the the ChexPert Dataset \cite{ChexPert}, which contains $224,316$ chest radiographs of $65,240$ patients as our second dataset. The initial trained model was trained on the NIH ChestXray14 dataset, as described previously. We selected $1057$ patient images from the ChexPert dataset having pleural effusion. The dataset was split into training ($70\%$), validation ($10\%$) and test ($20\%$), at the patient level such that all images from one patient are in a single fold.

\begin{figure}[t]
\begin{tabular}{c}
\includegraphics[height=4.5cm, width=8.5cm]{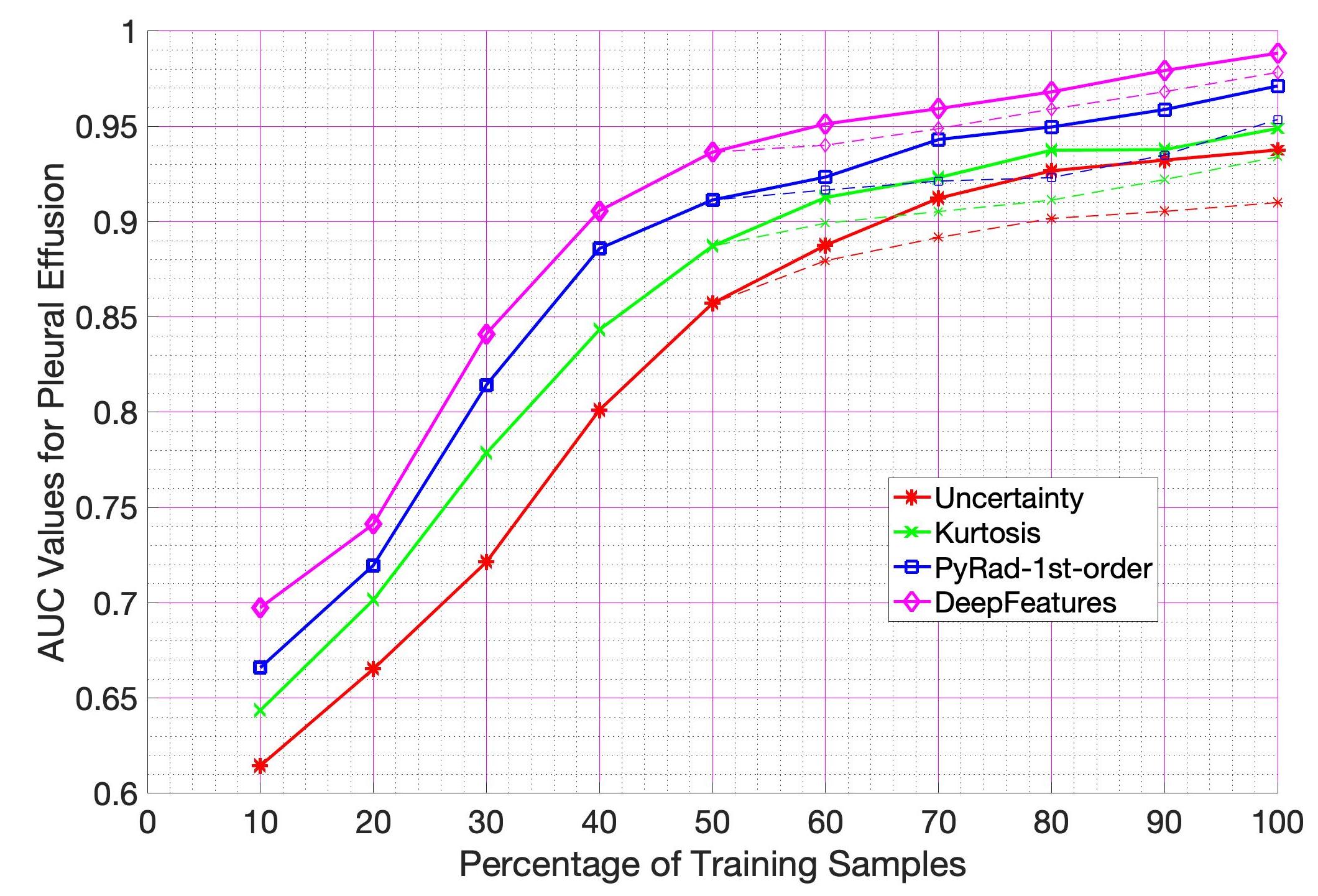} \\
(a) \\
\end{tabular}
\caption{Classification performance when switching between two datasets (NIH $\rightarrow$ ChexPert) at $50\%$. For comparison purposes dotted lines correspond to performance when no switching is performed.}
\label{fig:DatasetMix}
\end{figure}

To simulate this situation we started training with dataset 1 (e.g. NIH) and at mid way of training ($50\%$) we switched to dataset 2 (e.g. the CheXpert dataset \cite{ChexPert}). Figure~\ref{fig:DatasetMix}  shows the AUC curves for different IDEAL approaches. From this experiment we observe that the % following observation.
%\begin{enumerate}
    %\item In the initial stage (up to $50\%$ informative samples) we are using the NIH dataset. Hence the performance is similar to that observed in earlier plots of Figure~\ref{fig:ALcurves}. In this case the validation set is from the NIH dataset.
    %\item
    performance for both IDEAL and Uncertainty-based methods improved when switching to the CheXpert dataset, compared to the reference plots of Figure~\ref{fig:ALcurves} (shown as dotted lines in Figure~\ref{fig:DatasetMix}). The improvement can be attributed to the better quality of the CheXpert dataset (due to higher image resolution and higher SNR). This observation is also supported by the results in \cite{CheXNet} where the AUC values on the CheXpert dataset are higher than those reported for the NIH dataset for the same disease label. %We refrained from comparing both approaches since their regime of improvement lies in a different area of the AUC curve.

\subsection{Results For Semantic Segmentation}

For the task of histopathology image segmentation, the principle of informative sample selection holds true for segmentation as the segmentation network will benefit by learning from diverse and informative images. The motivation is to select the most informative images and their masks for training such that maximum performance gain can be achieved with minimal annotation cost. In order to derive saliency maps for the segmentation task, a classifier (DenseNet-121) for histopathology images (that identifies images as ``benign'' and ``malign'') was used as proxy to derive interpretability features and guide sample selection.

\begin{figure}[!htbp]
\begin{tabular}{c}
\includegraphics[height=4.9cm, width=7.99cm]{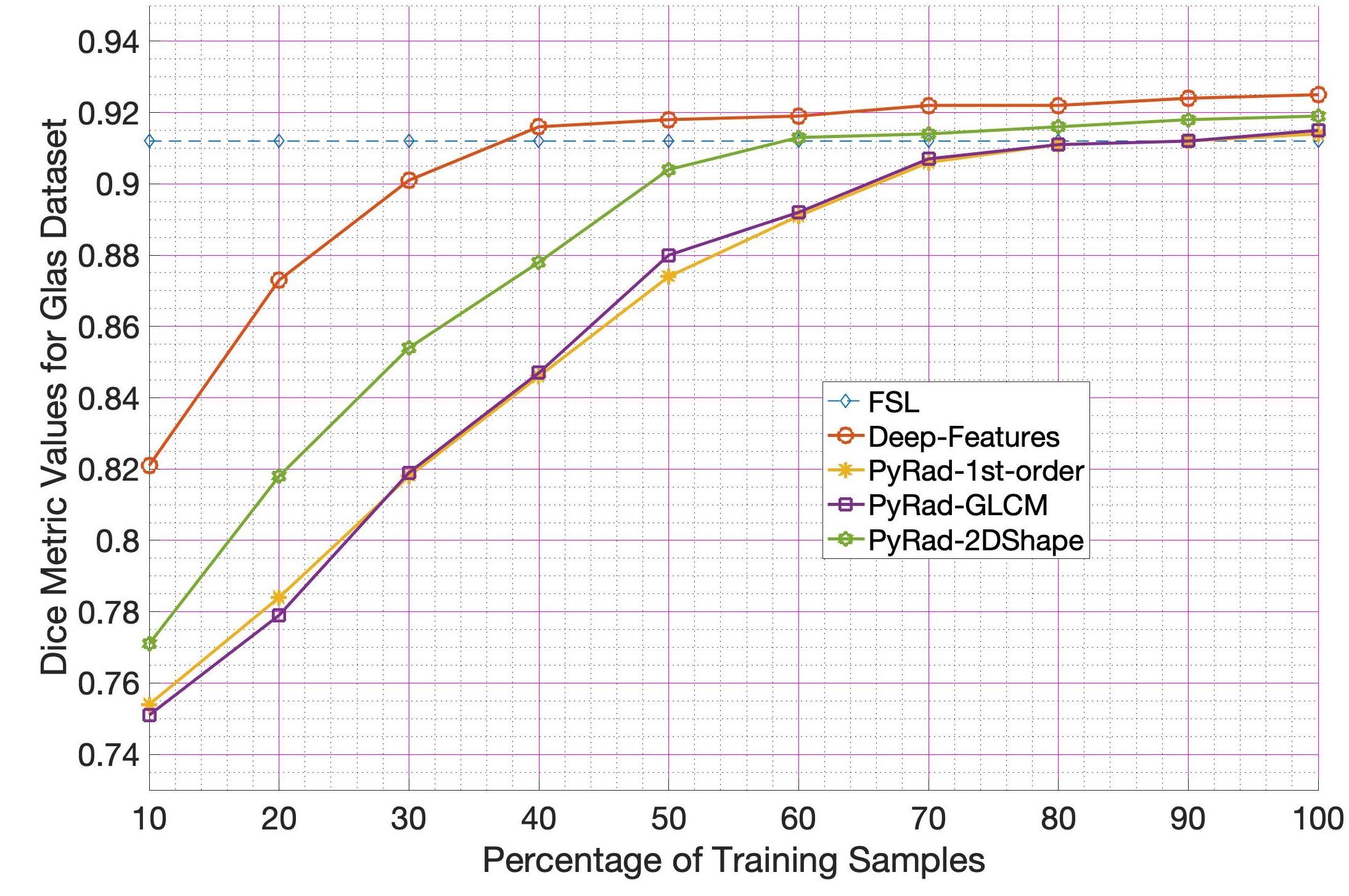} \\
(a) \\
\end{tabular}
\caption{Dice Metric values at different percentage levels of  training percentage for baselines and proposed IDEAL approach. IDEAL: Deep Saliency Features with proposed self-supervised ordinal clustering. }
\label{fig:segres}
\end{figure}

Similar to the approach for classification we identify informative samples based on classification labels and use the images with their masks for segmentation.
 A standard UNet \cite{Unet} is trained to perform segmentation, and the corresponding Dice metric values for every $10\%$ increase in dataset size are shown in Figure~\ref{fig:segres}. The UNet has $3$ convolution blocks followed by downsampling in the contracting path, followed by $3$ upsampling stages in the expansion stage. Each convolution block in the contracting path has $3$ convolution steps consisting of  $64, 3\times3$ filters with ReLU activation followed by batch normalization and $2\times2$ downsampling. In the expansion path each  deconvolution layer has stride $2$ followed by concatenation with the corresponding cropped feature map from the contracting path. It is followed by two $3 \times 3$ convolution layers with ReLU activation function (with batch normalization).

 In the initial stages when the segmentation network encounters new samples, the rate of increase of Dice metric is high but flattens in the later stages.  Similar to the  results for pleural effusion classification we also observed that sample selection outperforms fully supervised learning based segmentation.

\subsection{Performance on Simulated Noise}

\begin{figure}[!htbp]
\begin{tabular}{c}
\includegraphics[height=6.4cm, width=8.7cm]{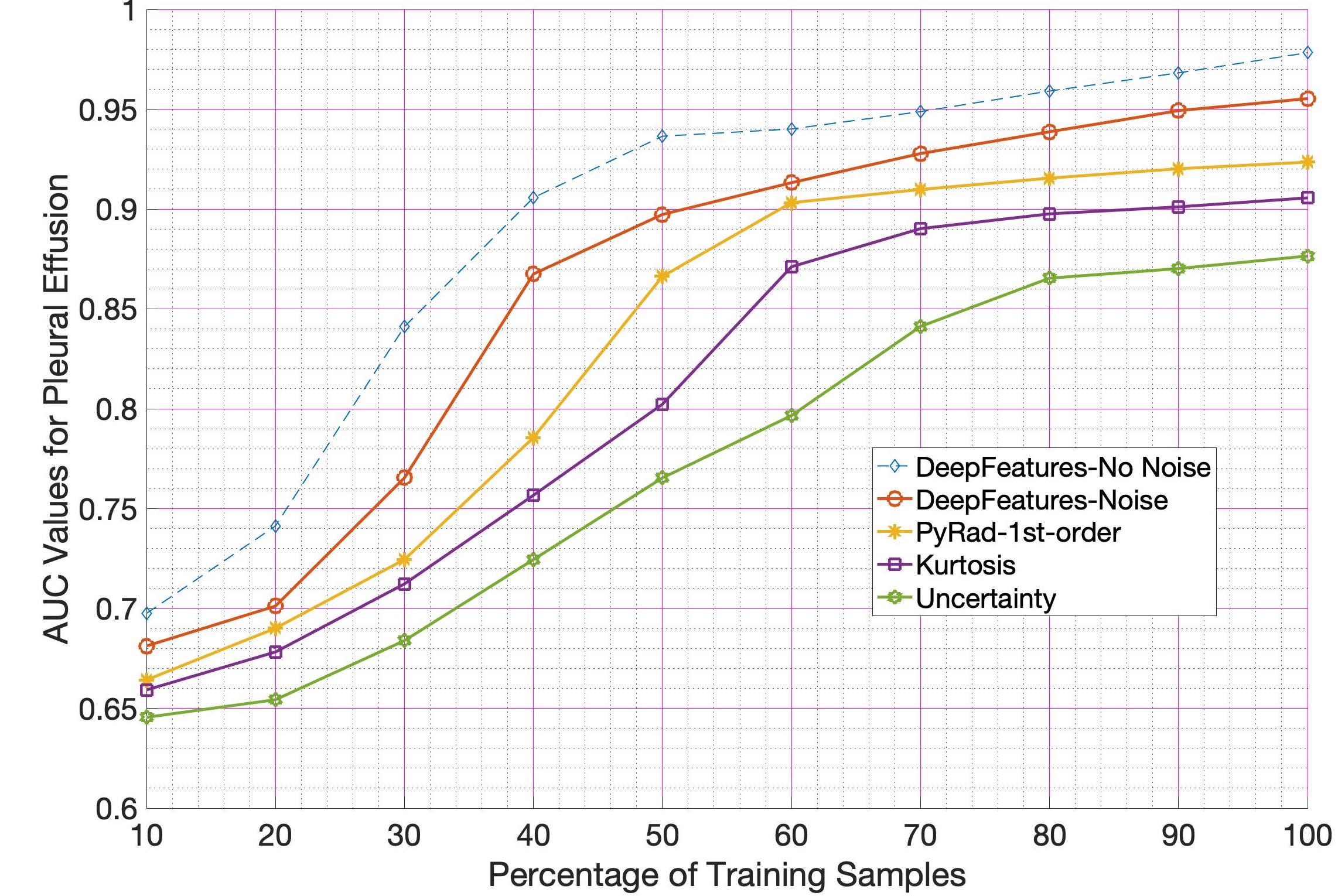} \\
\end{tabular}
\caption{AUC measures for different features for added Gaussian noise of $\mu=0,\sigma=0.05$. In dotted lines, `Deep Features' without noise are shown as reference. }
\label{fig:AL_noise}
\end{figure}

In an attempt to simulate low informativeness we added simulated noise of $\mu=0$ and different $\sigma\in \{0.005,0.01,0.05,0.1\}$.  Figure~\ref{fig:AL_noise} shows the AUC values for $\sigma=0.05$. The performance for `Deep-Features' without noise is shown as a dotted line for reference. With added noise the performance of all feature extraction methods degrade. However the deep features obtained using self supervision still perform the best and are more robust than other methods.

\subsection{Results on Pneumonia}

\begin{figure}[t]
\begin{tabular}{c}
\includegraphics[height=6.4cm, width=8.7cm]{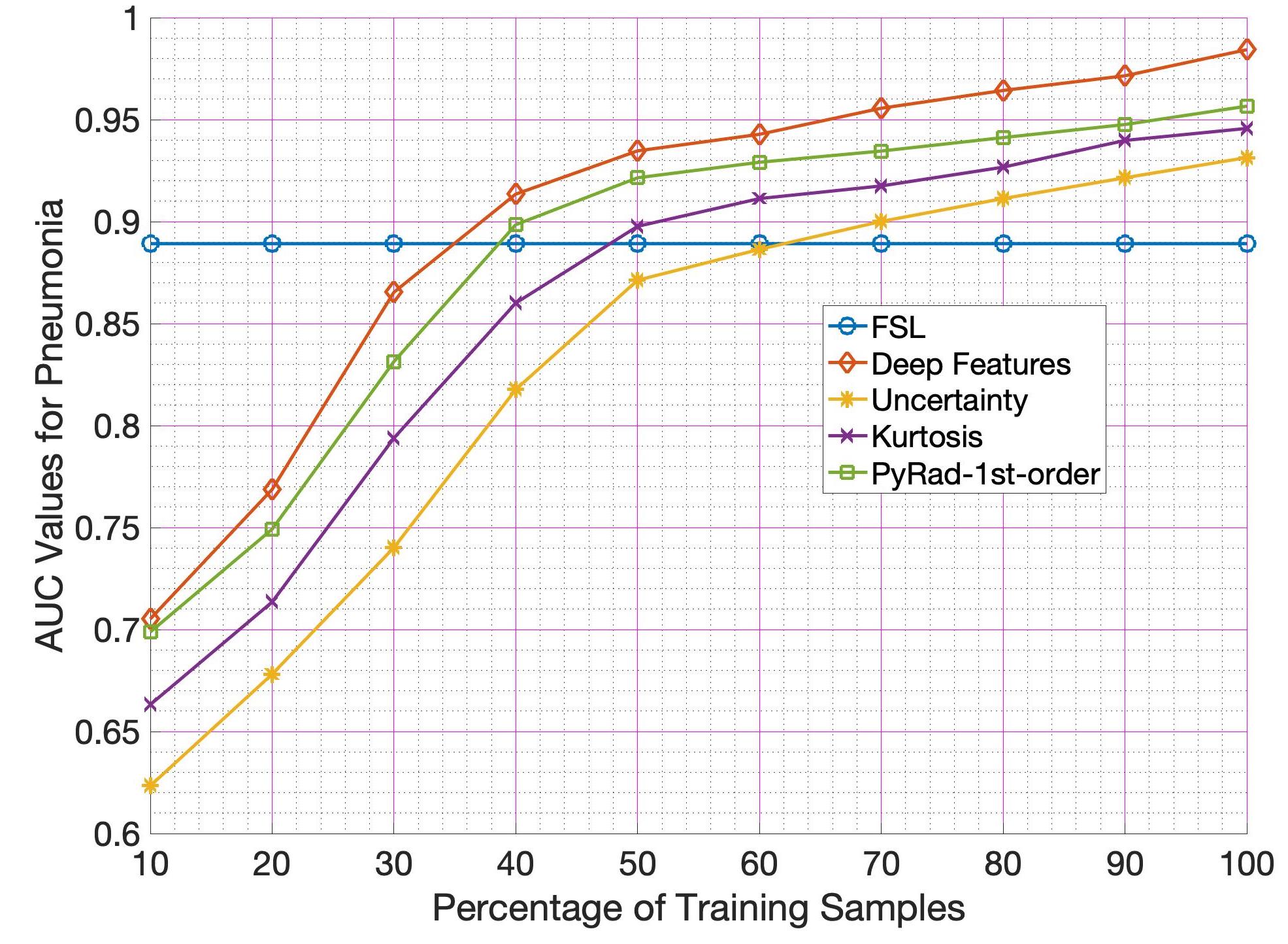} \\
\end{tabular}
\caption{Results on a second lung condition. AUC measures for Pneumonia at different percentage levels of  training percentage for baselines and proposed IDEAL approach. Plots are shown for Uncertainty, Kurtosis, Deep Saliency Features with proposed self-supervised ordinal clustering, and Radiomics 1st-order (PyRad-1st-order). As reference, AUC of a fully-supervised model (FSL) is also included as an horizontal line.}
\label{fig:ALcurves_pneumonia}
\end{figure}

We additionally tested on a second lung condition to check generalization of the findings on a different condition. Figure~\ref{fig:ALcurves_pneumonia} shows AUC plot for pneumonia. We used $1132$ images from different patients having pneumonia and the dataset  was  split  into  training  ($70\%$),  validation  ($10\%$)  and test ($20\%$), at the patient level such that all images from one patient are in a single fold. The characteristics of the different methods are similar to their pleural effusion counterpart in Figure~\ref{fig:ALcurves}. The results show that sample selection based on informativeness improves classifier performance for multiple diseases, and our proposed deep features do better than conventional feature extraction methods.

\subsection{Influence of Saliency Map Computation}

In this section we show results when using a different saliency map extraction method such as Grad-CAM \cite{gradcam} to check for generalization of the findings when using a different interpretability approach.

\begin{figure}[t]
\begin{tabular}{ccc}
\includegraphics[height=2.9cm, width=2.7cm]{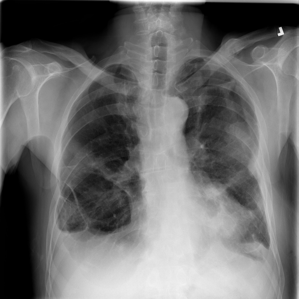} &
\includegraphics[height=2.9cm, width=2.7cm]{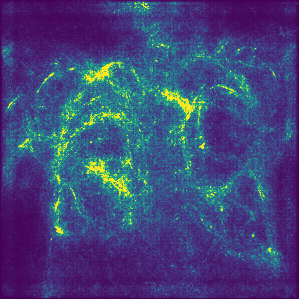} &
\includegraphics[height=2.9cm, width=2.7cm]{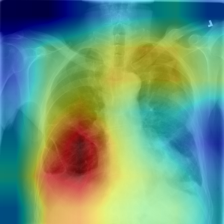} \\
\includegraphics[height=2.9cm, width=2.7cm]{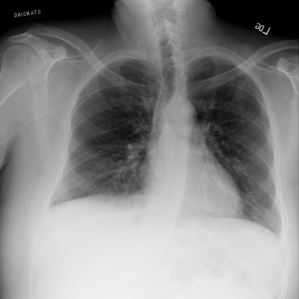} &
\includegraphics[height=2.9cm, width=2.7cm]{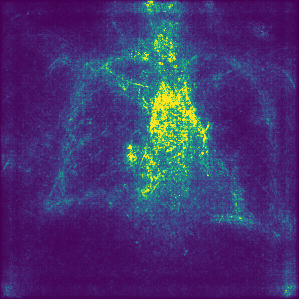} &
\includegraphics[height=2.9cm, width=2.7cm]{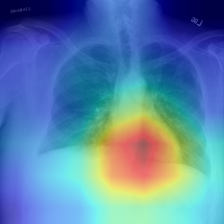} \\
(a) & (b) & (c) \\
\end{tabular}
\caption{Comparative visualization of GradCAM and Deep Taylor models. (a) original image; Saliency maps using (b) Deep Taylor method; (c) Grad-CAM method. Top row shows high informative image while bottom row shows a low informative image. Especially for low informative images, the Deep Taylor method gives a more accurate localization of informative regions than Grad-CAM.}
\label{fig:gcam}
\end{figure}

\begin{figure}[t]
\begin{tabular}{c}
\includegraphics[height=6.4cm, width=8.7cm]{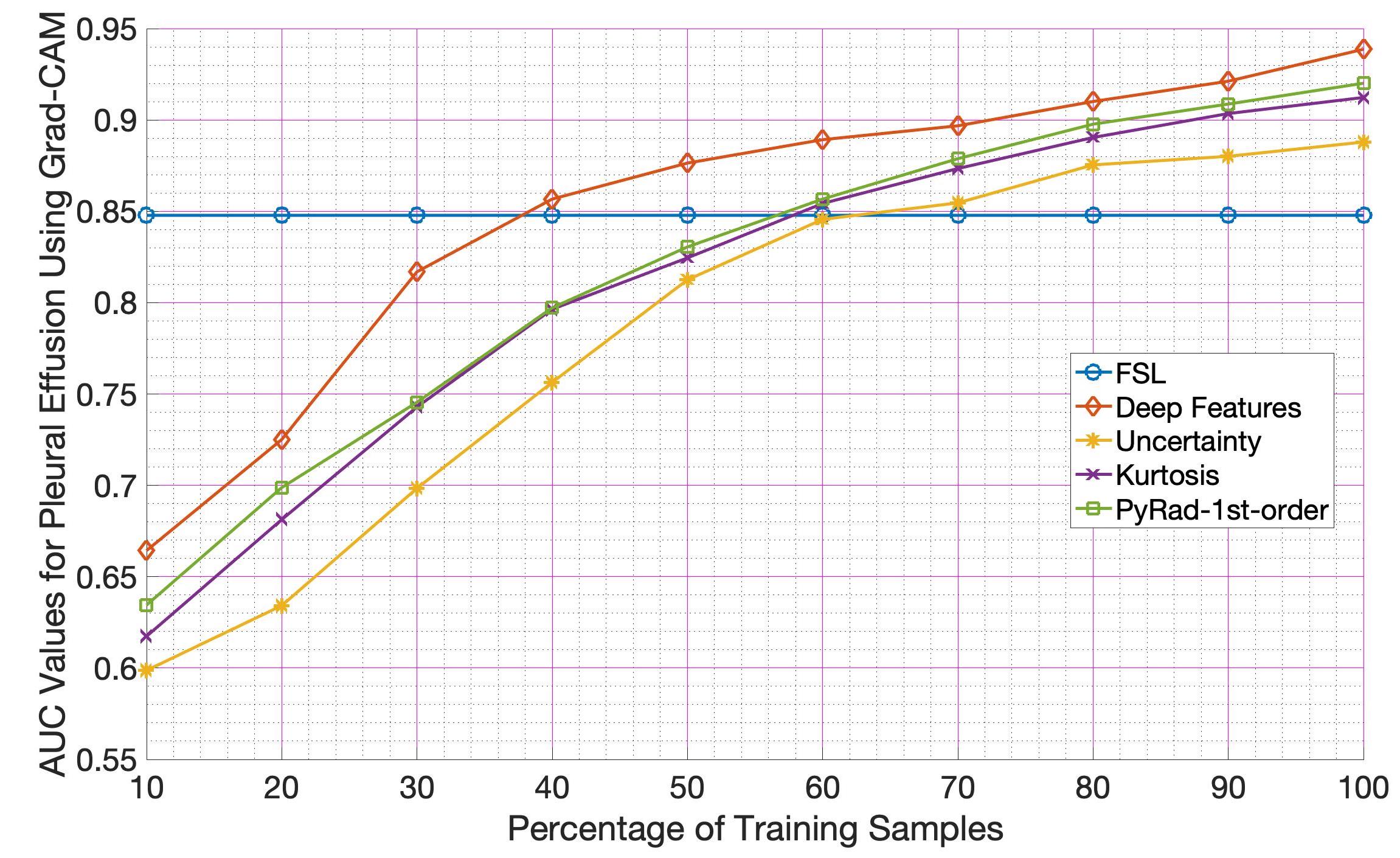} \\
\end{tabular}
\caption{AUC measures for Pleural Effusion using Grad-CAM saliency maps at different percentage levels of  training percentage for baselines and proposed IDEAL approach. Plots are shown for Uncertainty, Kurtosis, Deep Saliency Features with proposed self-supervised ordinal clustering, and Radiomics 1st-order (PyRad-1st-order). As reference, AUC of a fully-supervised model (FSL) is also included as an horizontal line.}
\label{fig:ALcurves_gradCAM}
\end{figure}

Figure~\ref{fig:gcam} shows the saliency map visualizations using Deep Taylor and Grad-CAM for high-informative and low-informative images. In the case of high-informative images  both approaches identify similar areas as salient. However, for low informative image (bottom row) the localized regions are quite different. Deep Taylor method highlights regions near the lung but the Grad-CAM method tend to localize an area beyond the lung region where there is no anatomy of interest. This justifies our choice of using Deep Taylor approach for generating saliency maps. Moreover, Figure~\ref{fig:ALcurves_gradCAM} shows AUC plots using Grad-CAM generated saliency maps. The trends are similar to Deep Taylor generated maps, showing the superiority of proposed approach over the baselines. However in comparison with Deep Taylor, the AUC values yield via Grad-CAM are lower for each of the corresponding feature extraction methods (see Figure~\ref{fig:ALcurves}). The plots also quantify the superior performance via Deep Taylor saliency maps, and point to the differences in interpretability maps studied in the literature \cite{doshivelez2017rigorous,Reyes2020}.

\subsection{Similarity Analysis Of Selected Images}

\subsubsection{Quality of Unlabeled Image Rankings}

We analyzed the rankings produced by the different sample selection approaches. To this end we used the normalized Discounted Cumulative Gain (nDCG) to compare rankings \cite{Fernandes2017}. 
The nDCG is defined as 
\begin{equation}
    nDCG_p=\frac{DCG_p}{IDCG_p},
\end{equation}
where DCG is Discounted Cumulative Gain and is defined as 
\begin{equation}
    DCG_p=\sum_{i=1}^{p} \frac{2^{rel_i}-1}{\log_2(i+1)},
\end{equation}
where $p$ represents the number of retrieved images considered. Relevance values ($rel_i$) were assigned from 1 to 5.5, 1 being the least similar image according to the reference ranking and 5.5 the most similar one (i.e., the relevance of two contiguous positions differs by 0.5).

\begin{figure}[t]
\begin{tabular}{c}
\includegraphics[height=4.4cm, width=6.7cm]{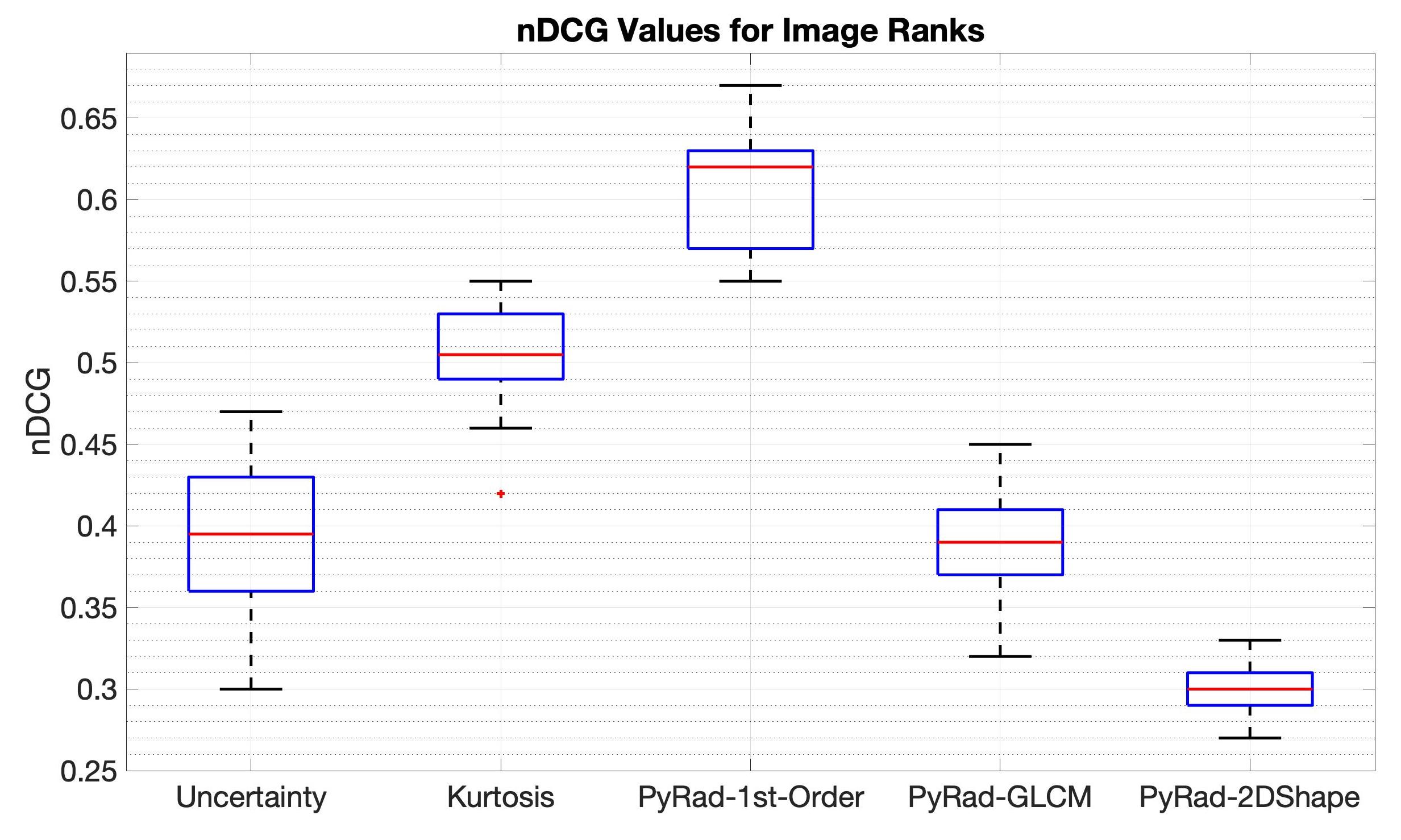} \\
\end{tabular}
\caption{Box plots of nDCG values of different methods compared with `Deep Features'. Higher values indicate better agreement with `Deep Features'.}
\label{fig:rank}
\end{figure}

The reference ranking was set as the Deep Features, which was compared with the informativeness ranking provided by the other methods. Figure~\ref{fig:rank} shows the comparison results, where higher values indicate better agreement. Particularly, the highest agreement was found between `Deep Features' and $PyRad_{1st-order}$ features. The plots support our previous observations (e.g., Figure~\ref{fig:ALcurves}) where $PyRad_{1st-order}$ showed to have the closest performance to `Deep Features'.

\subsubsection{Analysis of selected samples across methods}

\begin{figure}[t]
\begin{tabular}{c}
\includegraphics[height=6.4cm, width=8.7cm]{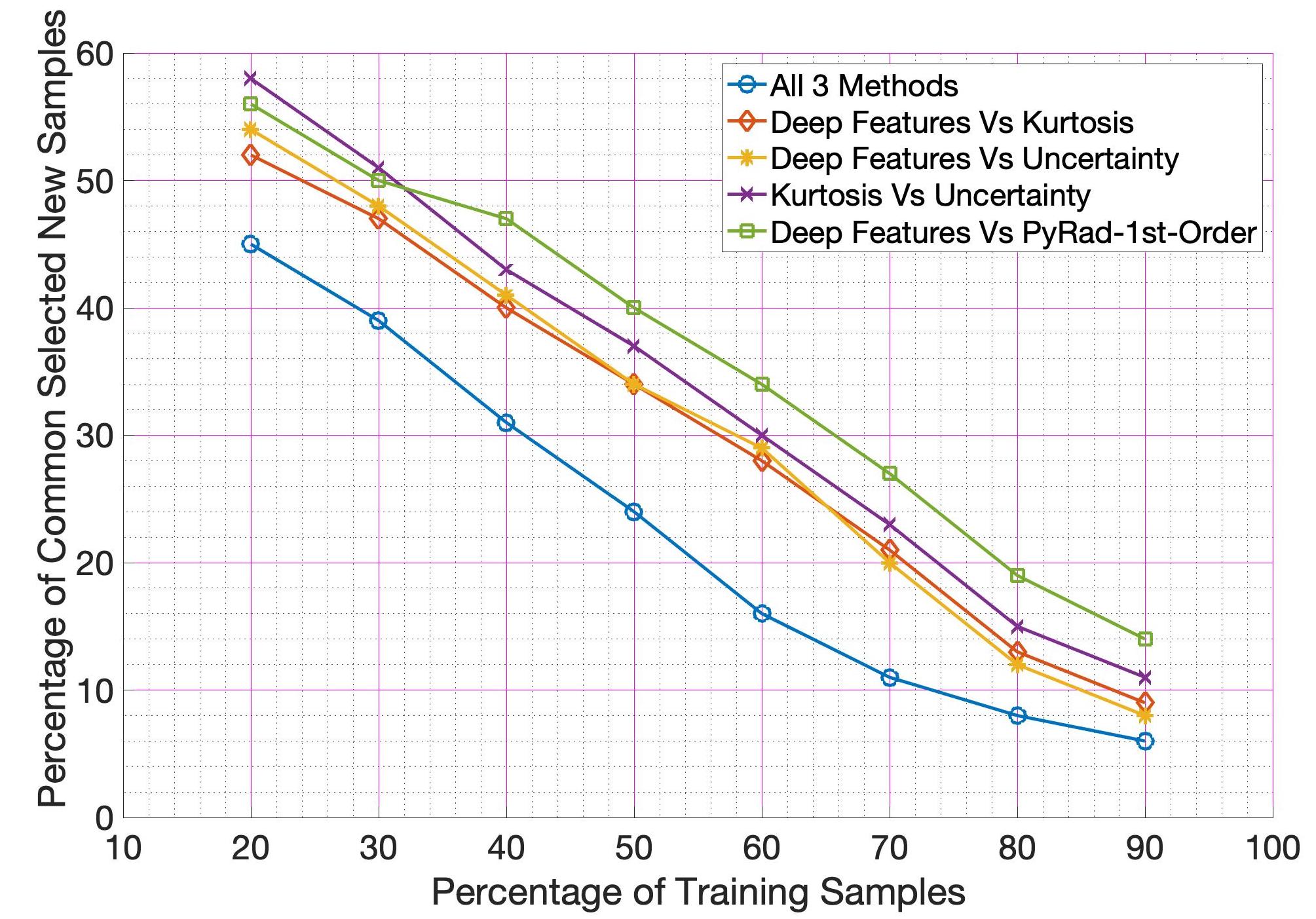} \\
\end{tabular}
% \caption{AUC measures at different percentage levels of training samples for different approaches.}
\caption{Percentage of common samples for every $10\%$ increase of training set. ``All 3 methods'' correspond to analyzing common samples when using Uncertainty, Kurtosis and Deep Features.}
\label{fig:corr}
\end{figure}

We further analyzed the number of common samples chosen by the different feature types of the IDEAL approach. Figure~\ref{fig:corr} shows the percentage of common samples chosen for the following informative sample selection approaches: 1) All $3$ methods - Kurtosis, PyRad$_{1st-Order}$ and Deep Features; 2) `Deep Features vs. Kurtosis'- the common informative samples chosen by our proposed Deep Features and Kurtosis; 3) `Deep Features vs. Uncertainty'- the common informative samples chosen by our proposed Deep Features and Uncertainty; 4) `Kurtosis vs. Uncertainty'- the common informative samples chosen by Kurtosis and Uncertainty; 5) `Deep Features vs. PyRad$_{1st-Order}$'- the common informative samples chosen by Deep Features and PyRad$_{1st-Order}$.
We observed from Figure~\ref{fig:corr} that during the initial stages the percentage of common samples is higher and then decreases subsequently. This is explained as in the initial stages there are more informative samples to choose from and hence a larger overlap of common samples appears. As the training progresses, the subsequent chosen informative samples tend to be different depending upon the accuracy of the classifier.

\subsection{Analysis Of Clinician Workload}

We engaged an experienced lung radiologist, with over 15 years of experience in analysing chest X-ray images, to assess differences when analyzing different levels of informative images, as selected by our method. We selected $100$ images having pleural effusion - split in two groups of $50$ high informative and $50$ low informative images. The level of informativeness was chosen by our algorithm using Deep Features. The clinician was blinded to the informativeness label as well as to the image's disease label, and was asked to diagnose the images by recording the time taken for each diagnosis.

On average the clinician spent $3.4$ seconds on each of the low informative images and he correctly diagnosed $36/50 (72\%)$ of the cases. For high informative images he spent on average $4.02$ seconds on each image and correctly diagnosed $41/50 (82\%)$ of the cases. The clinician spent an extra $18\%$ time (i.e. 0.62 seconds) time to diagnose the high informative images, and commented on the higher subtlety of the diagnosis as well as other factors such as fluid overload and not congestive heart failure, increasing the complexity. Hence, the time difference between diagnosing low and highly informative samples is negligible for an expert. The proposed approach requires $33\%$ of training samples to attain the same performance as a fully-supervised approach (Figure \ref{fig:ALcurves}(b)) compared to $50\%$ for the Uncertainty based baseline (Figure \ref{fig:ALcurves}(a). Our automated algorithm can classify an image in $0.2$ seconds. In light of these findings assessing clinician's workload and learning rates we conclude on the time benefits the proposed IDEAL approach can bring to clinicians. 
We also observed a slightly higher percentage of accurately diagnosed cases for high informative images. The higher information content of the images leads to a better diagnosis accuracy which is along expected lines. The results show that the images chosen as highly informative by our  method are indeed so as supported by the high analysis time and higher  diagnostic accuracy by clinicians. Thus our algorithm can effectively contribute to reducing clinician workload.

\section{Conclusions}

In this work we have presented results of an interpretability-driven active sample selection (IDEAL). IDEAL uses information from interpretability saliency maps to select informative samples for active learning. We propose a novel self supervised approach using deep features and ordinal clustering to determine the most informative sample. Results on publicly available datasets for lung pleural effusion and pneumonia disease classification and on histopathology image segmentation, show that the proposed IDEAL self supervised deep features outperform other methods in selecting the most informative samples for an effective active learning system. 
Additionally, the use of interpretability saliency maps provides experts with a mechanism to audit and monitor the active learning process, which we believe is an important added value of the proposed IDEAL approach.

As presented here, we believe interpretability approaches not only can be used to enhance understanding of model's predictions, but also to assist and provide further information of value in other areas of model performance, training and evaluation. In this regards, an area of potential research relates to the possibility of linking interpretability approaches with recent work on minimization of stochastic gradient to improve training of deep neural networks \cite{katharopoulos2018not,zhang2019autoassist}.

\section{Acknowledgements}
This work was supported by the Swiss National Foundation grant number 198388, and Innosuisse grant number 31274.1.

\appendix

\subsection{Training, Validation and Testing Loss Error Plots}

\begin{figure}[t]
\begin{tabular}{c}
\includegraphics[height=6.4cm, width=8.7cm]{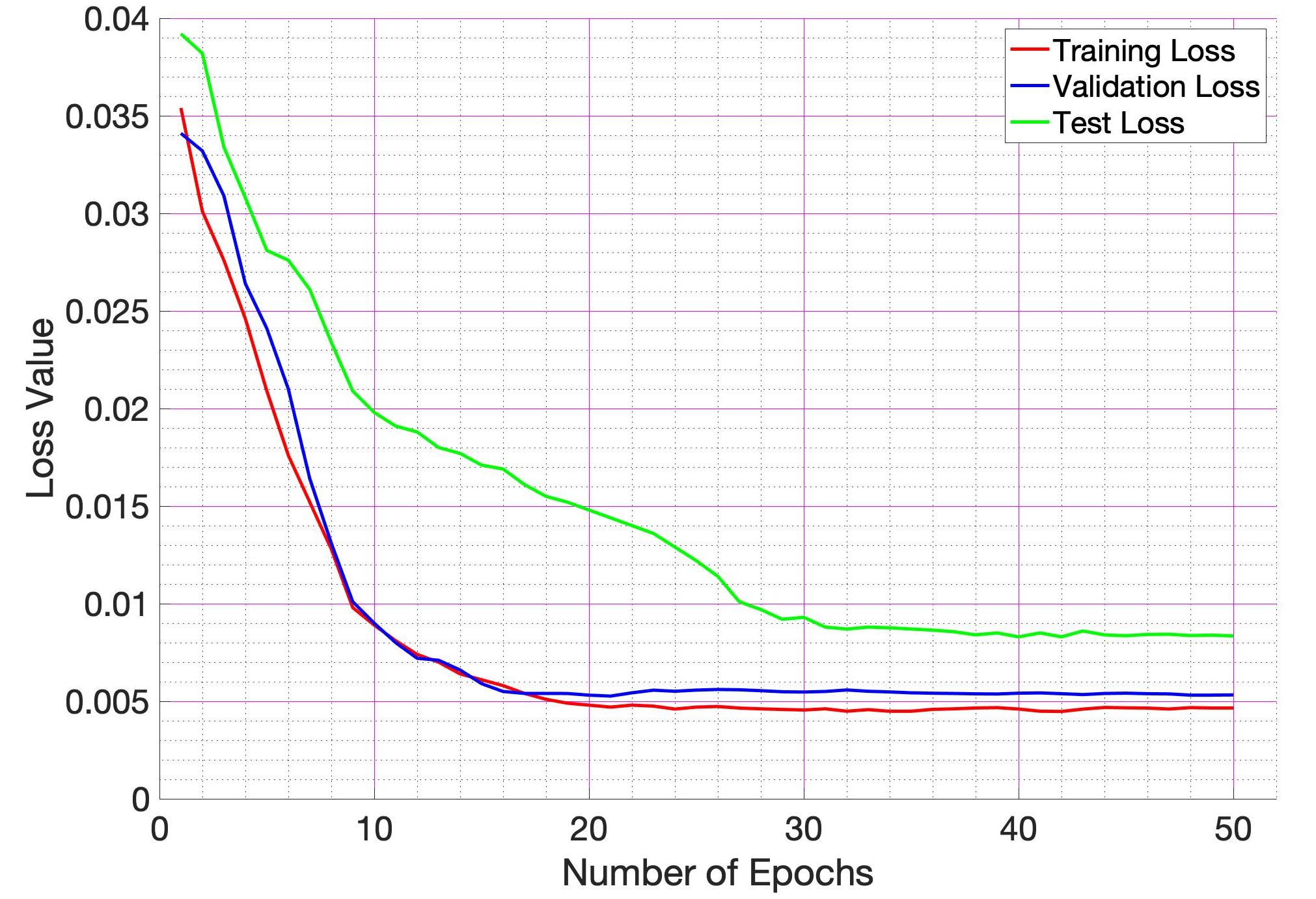} \\
\end{tabular}
\caption{Variation of Training, Validation and Test loss with increasing number of epochs. Values are for pleural effusion case.}
\label{fig:loss}
\end{figure}

Figure~\ref{fig:loss} shows the variation of training, validation and test error values with increasing number of epochs when training the model on pleural effusion images. The training and validation losses are at similar values indicating there is no overfitting to the training set. The test error is expectedly higher but not significantly when compared to the training loss. 

\bibliographystyle{IEEEtran}
\bibliography{ms}

\end{document}